 \documentclass[preprint,12pt]{elsarticle}

\makeatletter
\def\ps@pprintTitle{%
 \let\@oddhead\@empty
 \let\@evenhead\@empty
 \def\@oddfoot{}%
 \let\@evenfoot\@oddfoot}
\makeatother

\usepackage{graphicx}
\usepackage{epsfig}
\usepackage{xcolor}
\usepackage{amssymb}
\usepackage{amsmath}
\usepackage{multirow}
\usepackage{textcomp}
\usepackage{gensymb}
\usepackage{url}
\usepackage{tikz} 
\usepackage{xspace}

\biboptions{sort&compress}
\long\def\comment#1{} 

\newcommand{\ie}[0]{{\em i.e.}\ }
\newcommand{\etal}[0]{{\em et al.}\xspace }
\newcommand{\etc}[0]{{\em etc}\xspace}
\newcommand{\vs}[0]{{\em vs.}\xspace}
\newcommand{\vv}[0]{{\em vice-versa}\xspace}

\begin{document}
\begin{frontmatter}
\title{A model of figure ground organization incorporating local and global cues} 
\author{Sudarshan Ramenahalli}
\address{Department of Electrical and Computer Engineering, Johns Hopkins University, Baltimore, MD \\ sramena1@jhu.edu, sudarshan.rg@gmail.com}



\begin{abstract}
Figure Ground Organization (FGO) - inferring spatial depth ordering of objects 
in a visual scene - involves determining which side of an occlusion boundary is 
figure (closer to the observer) and which is ground (further away from the 
observer). A combination of global cues, like convexity, and local cues, like 
T-junctions are involved in this process. We present a biologically motivated, 
feed forward computational model of FGO incorporating convexity, surroundedness, 
parallelism as global cues and \textit{spectral anisotropy} (SA), T-junctions as 
local cues. While SA is computed in a biologically plausible manner, the 
inclusion of T-Junctions is biologically motivated. The model consists of three 
independent feature channels, Color, Intensity and Orientation, but SA and 
T-Junctions are introduced only in the Orientation channel as these properties 
are specific to that feature of objects. We study the effect of adding each 
local cue independently and both of them simultaneously to the model with no 
local cues. We evaluate model performance based on figure-ground classification 
accuracy (FGCA) at every border location using the BSDS 300 figure-ground 
dataset. Each local cue, when added alone, gives statistically significant 
improvement in the FGCA of the model suggesting its usefulness as an independent 
FGO cue. The model with both local cues achieves higher FGCA than the models 
with individual cues, indicating SA and T-Junctions are not mutually 
contradictory. Compared to the model with no local cues, the feed-forward model 
with both local cues achieves $\geq 8.78$\% improvement in terms of FGCA.

\end{abstract}


\end{frontmatter}

\section{Introduction}
\label{sec:CraftNatural-Intro}
An important step in the visual processing hierarchy is putting together 
fragments of features into coherent objects and inferring the spatial 
relationship between them. The feature fragments can be based on color, 
orientation, texture, \etc. 
Grouping~\cite{wagemans2012century,wagemans2012century_Part2} refers to the 
mechanism by which the feature fragments are put together to form perceptual 
objects. Such objects in the real world may be isolated, fully occluding one 
another or partially occluding, depending on the observer's viewpoint. In the 
context of partially occluding objects, Figure-ground organization (FGO) refers 
to determining which side of an occlusion boundary is the occluder, closer to 
the observer, referred to as \emph{figure} and which side is the occluded, far 
away from the observer, termed as \emph{ground}. 

Gestalt psychologists have identified a variety of cues that mediate the process 
of FGO~\cite{Koffka35}. Based on the spatial extent of information integration, 
these cues can be classified into local and global cues. Global cues such as 
symmetry~\cite{Bahnsen1928129}, surroundedness~\cite{Palmer99}, and 
size~\cite{Fowlkes2007} of regions integrate information over a large spatial 
extent to determine figure/ground relationship between objects. Local cues, on 
the other hand, achieve the same by analysis of only a small neighborhood near 
the boundary of an object. Hence, they are particularly attractive from a 
computational standpoint. Some examples of local cues are 
T-junctions~\cite{Heitger1992} and shading~\cite{Huggins_etal01}, including 
extremal edges~\cite{Palmer2008a,Ramenahalli2011a}. 

The neural mechanism by which FGO is achieved in the visual cortex is an active 
area of research, referred to as Border Ownership (BO) coding. The contour 
fragments forming an object's boundary are detected by Simple and Complex cells 
in the area V1 of primate visual cortex with their highly localized, 
retinotopically organized receptive fields. Cells in area V2, which receive 
input from V1 Complex cells, were found to code for BO by preferentially firing 
at a higher rate when the figural object was located on the preferred side of 
the BO coding neuron at its preferred orientation, irrespective of local 
contrast~\cite{Zhou_etal00}. Recently, \citet{williford2016figure}, remarkably 
show for the first time, that V2 neurons maintain the same BO preference 
properties even for objects in complex natural scenes. 

Many computational models~\cite{Craft2007,roelfsema2002figure,Zhaoping05} have 
been proposed to explain the neural mechanism by which FGO or BO coding is 
achieved in the visual cortex. Based on the connection mechanism, those models 
can be classified as feed-forward, 
feedback~\cite{Mihalas_etal11b, HuENEURO.0479-18.2019} or lateral interaction models~\cite{Zhaoping05}. 
In this work, we present a neurally motivated, feed-forward computational model 
of FGO incorporating both local and global cues. While we do not attempt to 
exactly mimic the neural processing at every step, we attempt to keep it as 
biologically motivated as possible. 

The FGO model we develop has three independent 
feature channels, Color, Intensity and Orientation. The main 
computational construct of the model is a BO computation mechanism that embodies 
Gestalt principles of convexity, surroundedness and parallelism, which is 
identical to all feature channels. In addition, we introduce many additional modifications 
to make it suitable for performing FGO and to incorporate local cues, as 
detailed in Section~\ref{sec:FGO-Model}. The model, applicable to any natural 
image, is tested on the widely used BSDS figure/ground dataset. First, we show 
that even the model with only global cues, devoid of any local cues achieves 
good results on the BSDS figure/ground dataset. Let us call this the 
\textit{Reference model}, against which we compare the performance of models with added 
local cues.

We add two local cues to the reference model, Spectral 
Anisotropy~\cite{ramenahalli2014local} and T-Junctions. The motivation behind 
adding local cues is their relatively low computational cost compared to global 
cues. Spectral Anisotropy (SA) was shown to be a valid cue for 
FGO~\cite{ramenahalli2014local,Ramenahalli2011a,Ramenahalli2012a} 
in predicting which side of an occlusion boundary is figure and 
which the background. Moreover, SA can be computed efficiently in a biologically 
plausible (See Section~\ref{subsec:SAcomputation}) manner using convolutions, 
making it an attractive candidate. T-Junctions are 
commonly viewed as one of the strongest cues of occlusion and 
their computation can be explained on the basis of end-stopped 
cells~\cite{Heitger1992,Heitger_vonderHeydt93,hansen2002biologically}. This is 
the biological motivation to incorporate them into the model. 

We have only a few FGO cues, specifically two local cues in our model. Both local cues 
influence the Orientation channel only as the properties they capture are more 
closely related to this feature. Certainly, many more local cues and global cues 
would be needed for best performance in real world images. But, here our primary 
 motivation is to develop a common computational framework and investigate how these local and global cues can be 
incorporated into a model of FGO. Second, our purpose is to verify whether 
local cues can co-exist along with the global cues. If so, how useful are these 
local cues? Can they lead to a statistically significant improvement in the 
model's performance when added alone? Finally, are these local cues mutually 
facilitatory leading to even further improvement, when added together? For these 
purposes, the minimalistic model with few global cues and even fewer local cues 
added to only one of the three feature channels provides an excellent analysis 
framework. Our goal is to study, from first principles, the effect of local and global 
cues in FGO, not necessarily to build a model with best performance. However, we compare the performance of our model with state of the art 
models of FGO, which are not biologically motivated, and show that our model performs competitively.

\section{Related Work}
\label{sec:CraftNatural-RelatedWork}
FGO has been an active area of research in Psychology since nearly a 
century~\cite{Rubin1921} ago. The Gestalt principles of FGO and grouping such as 
common fate, symmetry, good continuation, similarity \etc were formulated by Max 
Wertheimer~\cite{Wertheimer23} along with Kurt Koffka~\cite{Koffka35} and many 
others. Excellent reviews about the Gestalt principles of FGO and grouping can 
be found in \cite{wagemans2012century,wagemans2012century_Part2}. It is an 
active area of research in 
neuroscience~\cite{lamme1995neurophysiology,super2007altered,Zhou_etal00,williford2014early} 
and computer vision~\cite{ren2006figure,hoiem2011recovering,teo2015fast} as 
well. We limit our literature review to computational models only. Even though 
the terms ``FGO'', ``BO'' or ``grouping'' are not used in many publications we 
reviewed, the common goal in all of them is related to inferring depth ordering 
of objects. 

A local shapeme based model employing Conditional Random Fields (CRF) to enforce 
global consistency at T-junctions was proposed in~\cite{ren2006figure}. Hoiem 
\etal~\cite{hoiem2007recovering,hoiem2011recovering} used a variety of local 
region, boundary, Gestalt and depth based cues in a CRF model to enforce 
consistency between boundary and surface labels. An optimization framework~\cite{AmerRT10} to obtain a 2.1D 
sketch by constraining the ``hat'' of the T-junction to be figure and ``stem'' 
to be ground was proposed, which uses human labeled contours and T-junctions. In 
an extension~\cite{amer2014monocular}, a reformulated optimization over regions, 
instead of pixels, was proposed. By using various cues such as, curve and 
junction potentials, convexity, lower-region, fold/cut and parallelism,  
\citet{leichter2009boundary} train a CRF model to enforce global consistency. In a series of papers 
\citet{palou2013monocular,palou2012local,palou2011occlusion} show how image 
segmentation and depth ordering (FGO) can be performed using only low-level 
cues. Their model uses Binary Partition Trees (BPT)~\cite{salembier2000binary} 
for hierarchically representing regions of an image, performs depth ordering by 
iteratively pruning the branches of BPT enforcing constraints based on 
T-junctions and other depth related cues. In a recent work~\cite{teo2015fast}, 
which uses Structured Random Forests (SRF) for boundary detection, simultaneous 
boundary detection and figure-ground labeling is performed. They use shape cues, 
extremal edge basis functions~\cite{Ramenahalli2011a}, closure, image 
torque~\cite{NishigakiFD12} \etc to train the SRFs.  
 
\citet{Yu2001} present a hierarchical Markov Random Field (MRF) model 
incorporating rules for continuity of depth on surfaces, discontinuity at edges 
between surfaces and local cues such as T- and L-junctions. The model learns 
from a couple examples and effectively does depth segregation, thereby FGO. 
In~\cite{baek2005inferring}, a neurally plausible model integrating multiple 
figure-ground cues using belief propagation in Bayesian networks with leaky 
integrate and fire neurons was proposed. A simultaneous segmentation and 
figure-ground labeling algorithm was reported in \cite{maire2010simultaneous} 
which uses Angular Embedding~\cite{yu2009angular} to influence segmentation cues 
from figure-ground cues and \vv. Similar attempts with primary goal of 
segmenting images and labeling object classes using figure-ground cues can be 
seen in \cite{ion2011image,ion2014probabilistic}. 

Differentiation/Integration for Surface Completion (DISC) 
model~\cite{kogo2010surface} was proposed in which BO is computed by detecting 
local occlusion cues such as T- and L- junctions and comparing non-junction 
border locations with junction locations for BO consistency with the local cues. 
A Bayesian belief network based model was 
proposed~\cite{msingh_feldman_2010_bayesian} in which local cues (curvature and 
T-junctions) interact with medial axis or skeleton of the shape to determine BO.

In one of the early attempts~\cite{kienker1986separating}, a two layer network 
with connections between ``computational units'' within and across layers is 
proposed. These units integrate bottom-up edge input with top-down attention 
input to realize FGO. \citet{grossberg1985neural,grossberg1994} propose that a reciprocal interaction 
between a Boundary Contour System (BCS) extracting edges and a Feature Contour 
System (FCS) extracting surfaces achieves not only FGO, but also 3D perception. A model of contour grouping and FGO was proposed in 
\cite{Heitger_vonderHeydt93} central to which is a ``grouping'' mechanism. The 
model not only generates figure-ground labels, but also simulates the perception 
of illusory contours. Another influential model was proposed in 
\cite{Sajda_Finkel95} with feedback and feed-forward connections having 8 
different computational modules to obtain representations of contours, surfaces 
and depth.  \citet{roelfsema2002figure,jehee2007boundary} propose multilayer 
feedback networks resembling the neural connection pattern in the visual cortex 
to perform BO assignment through feedback from higher areas. Li Zhaoping 
\etal~\cite{zhaoping2003v1,li2000can} propose a model of FGO based on V1 
mechanisms. The model consists of orientation selective V1 neurons which 
influence surrounding neurons through mono-synaptic excitatory and di-synaptic 
suppressive connections. The excitatory lateral connections mimic colinear 
excitation~\cite{KAPADIA1995843} and cross-orientation 
facilitation~\cite{slllito1995visual}, while inhibitory connections model the 
iso-orientation suppression~\cite{Knierim_vanEssen92}. In a related 
model~\cite{Zhaoping05}, neurons in V2 having properties of convexity 
preference, good continuation and proximity was presented. A BO coding model 
which detects curvatures, L-Junctions and sends proportional signals  to a BO 
layer was proposed by \citet{kikuchi2001model}, where BO signals are propagated 
along the contour for two sides of BO. 

The model proposed by \citet{Craft2007} 
consists of edge selective cells, BO cells and multi-scale grouping (G) cells. 
The G cells send excitatory feedback to those BO cells that are co-circular and 
point to the center of the annular G cell receptive field. The model 
incorporates Gestalt principles of convexity, proximity and closure. 
But, it is a feedback model tested only on simple geometric shapes, not real-world natural images. Several 
models~\cite{Mihalas_etal11b,Russell_etal14,molin2013proto,Hu2017} 
with similar computational mechanisms have been proposed to explain various 
phenomena related to FGO, saliency, spatial attention, \etc. A model akin to 
\cite{Craft2007} was proposed in \cite{layton2012dynamic}, where in addition to 
G cells the model consists of region cells at multiple scales. In a feedback 
model~\cite{domijan2008feedback} based on the interaction between dorsal and 
ventral streams, surfaces which are of smaller size, greater contrast, convex, 
closed, having higher spatial frequency are preferentially determined as 
figures. The model also accounts for figure-ground cues such as lower region and 
top-bottom polarity. In a series of 
papers~\cite{sakai2012consistent,nishimura2004determination,nishimura2005computational} 
Sakai and colleagues formulate a BO model in which localized, asymmetric 
surround modulation is used to detect contrast imbalance, which then leads to 
FGO. \citet{russell2014model} propose a feed-forward model with Grouping and Border Ownership 
cells to study proto-object based saliency. Though our model is inspired by this work, 
the goal of \citet{russell2014model} model is to explain the formation of proto-objects~\cite{rensink2000dynamic} and saliency 
prediction, not Figure-Ground Organization. Another related model is proposed by \citet{HuENEURO.0479-18.2019},  
which is a recurrent model with feedback connections, devoid of any local cues. To the best of our 
knowledge our's is the first feed-forward model of FGO with both local and global cues. 
Also, this the the first such model tested on real-world images of the BSDS300 
figure-ground dataset commonly used as a benchmark for FGO in natural images.

\section{Model Description}
\label{sec:FGO-Model}
\begin{figure}
	\centering
	\includegraphics[width=1.0\textwidth]{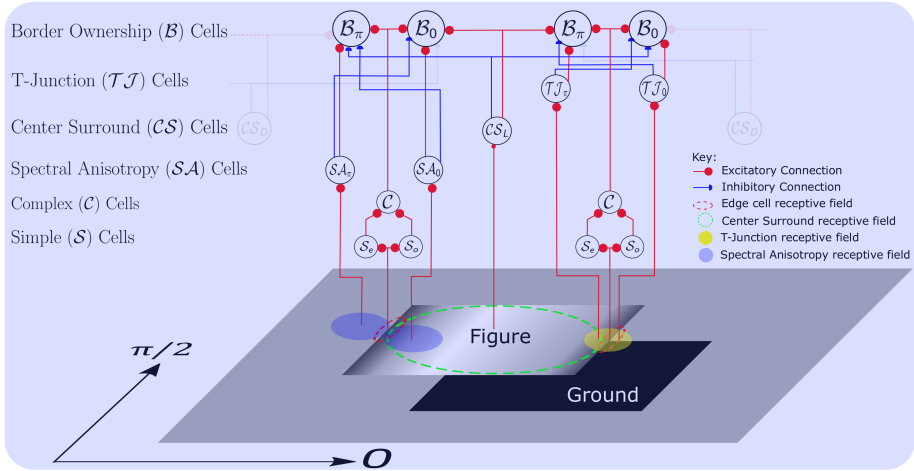}
	\caption{Figure-Ground Organization model with local cues: Input to the model are two 
overlapping squares. Bright foreground square has intensity gradient along the 
border (vertical orientation), which partially overlaps the black square forming 
T-Junctions. Network architecture for a single scale and single orientation, 
$\theta = \frac{\pi}{2}$ shown, but it is same for all 10 scales and 8 
orientations. Spectral Anisotropy ($\mathcal{SA}_{\theta \pm \frac{\pi}{2}}$) 
and T-Junctions ($\mathcal{TJ}_{\theta \pm \frac{\pi}{2}}$) are the local 
cues. $\mathcal{SA}$ and 
$\mathcal{TJ}$ cells are active only for the Orientation feature channel, as 
these are properties related only to that feature. Both $\mathcal{SA}$ and 
$\mathcal{TJ}$ cells excite $\mathcal{B}$ cells on the same side of the border 
and inhibit on the opposite side. $\mathcal{TJ}$ cue is computed such that 
$\mathcal{TJ}$ cells pointing to ``stem'' of T-Junction are zero, but have a 
high value for the opposite BO direction. 
}
	\label{fig:Paper2_Grouping_Illus}
\end{figure}

The model consists of three independent features channels, Color, Intensity and 
Orientation. The features are computed at multiple scales to achieve scale 
invariance. Orientation selective V1 Simple and Complex 
cells~\cite{Adelson_Bergen85} are excited by edge fragments of objects within 
their receptive field (Figure~\ref{fig:Paper2_Grouping_Illus}). Let us denote 
the contrast invariant response of a Complex cell at location $(x,y)$ by 
$\mathcal{C}_{\theta}(x,y,\omega)$, where $\theta$ is the preferred orientation 
of the cell and $\omega$ is the spatial frequency. As the spatial frequency 
(see Table~\ref{tab:Modelparams} for all parameters of the 
model) is same of all edge responsive cells in our model, except when explicitly 
stated otherwise (Section~\ref{subsec:SAcomputation}), we omit this variable for the most part. Each active Complex cell, 
$\mathcal{C}_{\theta}(x,y)$ activates a pair of BO cells, one with a BO 
preference direction, $\theta + \frac{\pi}{2}$ (a $90^{\circ}$ counter-clockwise 
rotation with respect to $\theta$) denoted as $\mathcal{B}_{\theta + 
\frac{\pi}{2}}(x,y)$, and the other with $\theta - \frac{\pi}{2}$ BO preference, 
denoted as $\mathcal{B}_{\theta - \frac{\pi}{2}}(x,y)$. When we talk about the 
BO response related to a specific figure/ground cue, be it local or global, a 
subscript is added to the right of the variable. For example, 
$\mathcal{B}_{\theta - \frac{\pi}{2},TJ}(x,y)$ would be used to denote the BO 
response related to T-Junctions. Likewise, when specifying scale is necessary, 
it is denoted by superscript, $k$. For example, $\mathcal{C}_{\theta}^{k}(x,y)$ 
denotes Complex cell response for orientation $\theta$ at location, $(x,y)$ and 
scale, $k$. On the other hand, when we need to explicitly specify the feature we 
talk about, a subscript is added to the left of the variable. For example, 
${}_{C}\mathcal{B}_{\theta - \frac{\pi}{2}}(x,y)$ represents the BO response for 
the Color feature channel. When a specific BO direction, feature, cue, scale or 
a location is not important, we just refer to them as, $\mathcal{B}$ cells, 
$\mathcal{C}$ cells, \etc. Same applies in all such situations. 

Without the influence of any local or global cues, the responses of both BO 
cells at a location will be equal, hence the figure direction at that location 
is arbitrary. The center-surround cells, denoted as $\mathcal{CS}$ cells, bring 
about global scene context integration by modulating the $\mathcal{B}$ cell 
activity. The $\mathcal{CS}_{L}$ cells (Figure~\ref{fig:Paper2_Grouping_Illus}) 
extract light objects on dark background, while $\mathcal{CS}_{D}$ cells code 
for dark objects on light background. Without the influence of local cues, this 
architecture embodies the Gestalt properties of convexity, surroundedness and 
parallelism (global cues). 

The local cues (see Section~\ref{sec:localCueComp} for computational details of 
local cues) modulate $\mathcal{B}$ cell activity additionally. Similar to 
$\mathcal{B}$ cells, a pair of Spectral Anisotropy cells exist for the two 
opposite BO preference directions at each location, which capture local texture 
and shading gradients (see Section~\ref{subsec:SAcomputation} for SA 
computation) on the two sides of the border. Let us denote by 
$\mathcal{SA}_{\theta + \frac{\pi}{2}}(x,y)$ the cell capturing Spectral 
Anisotropy for $\theta + \frac{\pi}{2}$ BO direction, likewise 
$\mathcal{SA}_{\theta - \frac{\pi}{2}}(x,y)$ for the opposite BO direction. The 
T-Junction cells (see Section~\ref{subsec:TJcomputation} for computational 
details) also come in pairs, for the two opposite BO directions. Similar to 
$\mathcal{SA}$ cells, $\mathcal{TJ}_{\theta \pm \frac{\pi}{2}}(x,y)$ hold the 
T-Junction cue information for the two antagonistic BO directions, $\theta \pm 
\frac{\pi}{2}$. Both these type of cells excite $\mathcal{B}$ cells of the same 
BO direction and inhibit the opposite BO direction $\mathcal{B}$ cells. For 
example, $\mathcal{SA}_{\theta + \frac{\pi}{2}}(x,y)$ excites 
$\mathcal{B}_{\theta + \frac{\pi}{2}}(x,y)$ and inhibits $\mathcal{B}_{\theta - 
\frac{\pi}{2}}(x,y)$. 

The influence of $\mathcal{CS}$ cells, $\mathcal{SA}$ cells and $\mathcal{TJ}$ 
cells on $\mathcal{B}$ cells is controlled by a set of weights (not shown in 
Figure~\ref{fig:Paper2_Grouping_Illus}). Local cues are active in the 
Orientation channel only. The interplay of all these cues leads to the emergence 
of figure/ground relations strongly biased for one of the two BO directions at 
each location. The network 
architecture depicted in Figure~\ref{fig:Paper2_Grouping_Illus} is the same 
computational construct that is applied at every scale, for every feature and 
orientation. The successive stages of model computation are explained in the 
following subsections. 

\subsection{Computation of feature channels}
\label{subsec:FGO-feat-comp}
We consider Color, Intensity and Orientation as three independent feature 
channels in our model, the computation of each is described in the following 
sections.

\subsubsection{Intensity channel}
\label{subsubsec:FGO-feat-comp-vis-I}
The input image consists of Red ($r$), Blue ($b$) and Green ($g$) color 
channels. The intensity channel, $I$ is computed as average of the three 
channels, $I = (r + b + g)/3$. As with all other feature channels, a 
multi-resolution image pyramid is constructed from the intensity channel 
(Section~\ref{subsec:FGO-multi-res-pyr}). The multi-resolution analysis allows 
incorporation of scale invariance into the model.

\subsubsection{Color opponency channels}
\label{subsubsec:FGO-feat-comp-vis-C}
The color channels are first normalized by dividing each $r$, $g$ or $b$ value 
by $I$. From the normalized $r$, $g$, $b$ channels, four color channels, Red 
($\mathcal{R}$), Green ($\mathcal{G}$), Blue ($\mathcal{B}$) and Yellow ($\mathcal{Y}$) are computed as,

\begin{equation}
\mathcal{R} = \max\Bigg(0,r - \frac{g + b}{2}\Bigg)
\label{eq:FGO-feat-comp-vis-RChannel}
\end{equation}

\begin{equation}
\mathcal{G} = \max\Bigg(0,g - \frac{r + b}{2}\Bigg)
\label{eq:FGO-feat-comp-vis-GChannel}
\end{equation}

\begin{equation}
\mathcal{B} = \max\Bigg(0,b - \frac{g + r}{2}\Bigg)
\label{eq:FGO-feat-comp-vis-BChannel}
\end{equation}

\begin{equation}
\mathcal{Y} = \max\Bigg(0,\frac{r + g}{2} - \frac{\left | (r - g) \right |}{2} - b\Bigg)
\label{eq:FGO-feat-comp-vis-YChannel}
\end{equation}
In Eq~\ref{eq:FGO-feat-comp-vis-YChannel}, the symbol, $\left | \quad \right |$ 
denotes absolute value. 

The four opponent color channels, $\mathcal{RG}$, $\mathcal{GR}$, $\mathcal{BY}$ 
and $\mathcal{YB}$ are computed as,
\begin{equation}
\mathcal{RG} = \max(0,\mathcal{R} - \mathcal{G})
\label{eq:FGO-feat-comp-vis-RGChannel}
\end{equation}

\begin{equation}
\mathcal{GR} = \max(0,\mathcal{G} - \mathcal{R})
\label{eq:FGO-feat-comp-vis-GRChannel}
\end{equation}

\begin{equation}
\mathcal{BY} = \max(0,\mathcal{B} - \mathcal{Y})
\label{eq:FGO-feat-comp-vis-BYChannel}
\end{equation}

\begin{equation}
\mathcal{YB} = \max(0,\mathcal{Y} - \mathcal{B})
\label{eq:FGO-feat-comp-vis-YBChannel}
\end{equation}

\subsubsection{Orientation channel}
\label{subsubsec:FGO-feat-comp-vis-O}
The Orientation channel is computed using the canonical model of visual 
cortex~\cite{Adelson_Bergen85}, where quadrature phase, orientation selective, 
Gabor kernels are used to model the V1 simple cells. The responses of Simple 
cells are non-linearly combined to obtain the contrast invariant, orientation 
selective response of the Complex cell. Mathematically, the receptive fields of 
even and odd symmetric Simple cells can be modeled as the cosine and sine 
components of a complex Gabor function - a sinusoidal carrier multiplied by a 
Gaussian envelope. The RF of a Simple Even cell, $s_{e,\theta}(x,y)$ is given 
by,

\begin{equation}
\label{eq:SimpleEven}
	s_{e,\theta}(x,y) = e^{-\frac{X^2 + \gamma^2 Y^2}{2 \sigma^2}}\cos(\omega X)
\end{equation}
where, $X = x\cos(\theta) + y\sin(\theta)$ and $Y = -x\sin(\theta) + 
y\cos(\theta)$ are the rotated coordinates, $\sigma$ is the standard deviation 
of the Gaussian envelope, $\gamma$ is the spatial aspect ratio (controlling how 
elongated or circular the filter profile is), $\omega$ is the spatial frequency 
of the cell and $\theta$ is the preferred orientation of the simple cell. 
Similarly, the receptive field of a Simple Odd cell is defined as,

\begin{equation}
\label{eq:SimpleOdd}
	s_{o,\theta}(x,y) = e^{-\frac{X^2 + \gamma^2 Y^2}{2 \sigma^2}}\sin(\omega X)
\end{equation}

Simple even and odd cells responses, respectively denoted $S_{e,\theta}(x,y)$ 
and $S_{o,\theta}(x,y)$ are computed by correlating the intensity image, 
$I(x,y)$ with the respective RF profiles. The Complex cell response, 
$\mathcal{C}_{\theta}(x,y)$ is calculated as,

\begin{equation}
\label{eq:ComplexCellResp}
\mathcal{C}_{\theta}(x,y) = \sqrt{S_{e,\theta}(x,y)^{2} + S_{o,\theta}(x,y)^{2}}
\end{equation}

Eight orientations in the range, $ [0,\pi] $, at intervals of $\frac{\pi}{8}$ are used. 
\subsection{Multiscale pyramid decomposition}
\label{subsec:FGO-multi-res-pyr}
Let us denote a feature map, be it Orientation ($\mathcal{C}_{\theta}$), Color 
($\mathcal{RG}$, $\mathcal{BY}$, \etc) or Intensity feature map, at image 
resolution by a common variable, $\beta^{0}(x,y)$. The next scale feature map, 
$\beta^{1}(x,y)$ is computed by downsampling $\beta^{0}(x,y)$. The downsampling 
factor can be either $\sqrt{2}$ (half-octave) or 2 (full octave). Bi-linear 
interpolation is used to compute values in the down-sampled feature map, 
$\beta^{1}(x,y)$, which is the same interpolation scheme used in all cases of 
up/down sampling. Similarly, any feature map $\beta^{k}(x,y)$ of a lower scale, 
$k$ is computed by downsampling the higher scale feature map, $\beta^{k-1}(x,y)$ 
by the appropriate downsampling factor. As the numerical value of $k$ increases, 
the resolution of the map at that level in the pyramid decreases. The feature 
pyramids thus obtained are used to compute BO pyramids explained the next 
section. 

In addition to the multiscale pyramids of independent feature channels, we compute the multiscale local cue pyramids for SA and 
T-Junctions as well. To denote the local cue map at a specific scale, as with 
feature pyramids, the scale parameter $k$ is used. For example, 
$\mathcal{SA}_{\theta + \frac{\pi}{2}}^{k}(x,y)$ denotes the Spectral Anisotropy 
feature map for $\theta + \frac{\pi}{2}$ border ownership direction at scale, 
$k$. Similarly T-Junction pyramids at different scales for $\theta \pm 
\frac{\pi}{2}$ BO directions are denoted by $\mathcal{TJ}_{\theta \pm 
\frac{\pi}{2}}^{k}(x,y)$. The local cue pyramids are computed by successively 
downsampling the local cue maps at native resolution, $\mathcal{SA}_{\theta \pm 
\frac{\pi}{2}}$ and $\mathcal{TJ}_{\theta \pm \frac{\pi}{2}}$ (see 
Section~\ref{sec:localCueComp} for their computation details).

\subsection{Border Ownership pyramid computation}
\label{subsec:FGO-multi-res-BO-pyr}
The operations performed on any of the features ($\mathcal{C}_{\theta}$ or $I$) 
or the sub-type of features like $\mathcal{RG},  \mathcal{BY}$ is the same. BO 
responses are computed by modulating $C_{\theta}(x,y)$ by the activity of 
center-surround feature differences on either sides of the border. Each feature map, $\beta^{k}(x,y)$, is correlated with 
the center-surround filters to get center-surround ($\mathcal{CS}$) difference 
feature pyramids. Two types of center-surround filters, $cs_{on}(x,y)$ 
(ON-center) and $cs_{off}(x,y)$ are defined as,
\begin{equation}
cs_{on}(x,y) = \frac{1}{2\pi\sigma_{in}^2} e^{-\frac{(x^2 + y^2)}{2\sigma_{in}^2}} - \frac{1}{2\pi\sigma_{out}^2} e^{-\frac{(x^2 + y^2)}{2\sigma_{out}^2}} 
\label{eq:FGO-cs-on-filt}
\end{equation}

\begin{equation}
cs_{off}(x,y) = -\frac{1}{2\pi\sigma_{in}^2} e^{-\frac{(x^2 + y^2)}{2\sigma_{in}^2}} + \frac{1}{2\pi\sigma_{out}^2} e^{-\frac{(x^2 + y^2)}{2\sigma_{out}^2}}
\label{eq:FGO-cs-off-filt}
\end{equation}
where $\sigma_{out}, \sigma_{in}$ are the standard deviations of the outer and 
inner Gaussian kernels respectively.

The center-surround dark pyramid, $\mathcal{CS}^{k}_{D}$ is obtained by 
correlating the feature maps, $\beta^k$ with the $cs_{off}(x,y)$ filter followed 
by half-wave rectification,
\begin{equation}
\mathcal{CS}^{k}_{D}(x,y) = \max(0,\beta^{k}(x,y) \ast cs_{off}(x,y))
\label{eq:FGO-cs-D-pyr}
\end{equation}
which detects weak/dark features surrounded by strong/light ones. In 
Eq~\ref{eq:FGO-cs-D-pyr}, the symbol, $\ast$ denotes 2D 
correlation~\cite{2Dcorr}. Similarly, to detect strong features surrounded by 
weak background, a $\mathcal{CS}^{k}_{L}$ pyramid is computed as,
\begin{equation}
\mathcal{CS}^{k}_{L}(x,y) = \max(0,\beta^{k}(x,y) \ast cs_{on}(x,y))
\label{eq:FGO-cs-L-pyr}
\end{equation}

The $\mathcal{CS}$ pyramid computation is performed this way for all feature 
channels except for the Orientation channel. For the Orientation feature 
channel, feature contrasts are not typically symmetric as in the case of other 
features, but oriented at a specific angle. Hence, the $cs_{on}(x,y)$ and 
$cs_{off}(x,y)$ filter kernels in Equations~\ref{eq:FGO-cs-D-pyr} and 
\ref{eq:FGO-cs-L-pyr} are replaced by even symmetric Gabor filters, 
$s_{e,\theta}(x,y)$ (ON-center) and $-s_{e,\theta}(x,y)$ (OFF-center) of 
opposite polarity respectively. But, in this case, different set of parameter 
values are used. Instead of $\gamma = 0.5$, $\sigma = 2.24$ and $\omega = 1.57$ 
used in Section~\ref{subsubsec:FGO-feat-comp-vis-O}, here we use $\gamma_{1} = 
0.8$, $\sigma_{1} = 3.2$ and $\omega_{1} = 0.7854$ respectively. The parameter 
values are modified in this case such that the width of the center lobe of the 
even Gabor filters (ON and OFF-center) matches the zero crossing diameter of the 
$cs_{on}(x,y)$ and $cs_{off}(x,y)$ filter kernels in 
Equations~\ref{eq:FGO-cs-D-pyr} and \ref{eq:FGO-cs-L-pyr}. As a result, the 
ON-center Gabor kernel detects bright oriented edges in a dark background, 
instead of symmetric feature discontinuities detected by $cs_{on}(x,y)$. 
Similarly, the OFF-center Gabor filter detects activity of dark edges on bright 
backgrounds. 

An important step in BO computation is normalization of the center-surround 
feature pyramids, $\mathcal{CS}^{k}_{L}(x,y)$ and $\mathcal{CS}^{k}_{D}(x,y)$. 
Let $\mathcal{N}(.)$ be used to denote the normalization operation, which is 
same as the normalization used in \cite{Itti1998}, but done after rescaling 
$\mathcal{CS}_{D}$ and $\mathcal{CS}_{L}$ pyramids to have the same range, $[0, 
\ldots, M]$. Similarly the local cue pyramids, $\mathcal{SA}_{\theta + 
\frac{\pi}{2}}^{k}(x,y)$ and $\mathcal{SA}_{\theta - \frac{\pi}{2}}^{k}(x,y)$ 
are also  normalized using the same method and in the same range, 
$[0,\ldots,M]$. In the same way, $\mathcal{TJ}_{\theta \pm 
\frac{\pi}{2}}^{k}(x,y)$ pyramids are also normalized. This normalization step 
enables comparison of different features and local cues on the same scale, hence 
the combination of feature and local cue pyramids. 

Since, we compute BO on the normalized light and dark CS pyramids, 
$\mathcal{N}(\mathcal{CS}_{L}^{k}(x,y))$ and 
$\mathcal{N}(\mathcal{CS}_{D}^{k}(x,y))$ separately and combine them at a later 
stage, let us denote, the corresponding BO pyramids by $B_{\theta \pm 
\frac{\pi}{2}, L}^{k}(x,y)$ and $B_{\theta \pm \frac{\pi}{2}, D}^{k}(x,y)$ 
respectively. We explain the BO pyramid computation for $B_{\theta + 
\frac{\pi}{2}, L}^{k}(x,y)$ and $B_{\theta + \frac{\pi}{2}, D}^{k}(x,y)$ which 
have a BO preference direction of $\theta + \frac{\pi}{2}$. Computation of 
$B_{\theta - \frac{\pi}{2}, L}^{k}(x,y)$ and $B_{\theta - \frac{\pi}{2}, 
D}^{k}(x,y)$ is analogous.

Let $\hat{K}_{\theta + \frac{\pi}{2}}(x,y)$ denote the kernel responsible for 
mapping the object activity from normalized $\mathcal{CS}_{L}$ and 
$\mathcal{CS}_{D}$ pyramids to the object edges, which is implemented with von 
Mises distribution. von Mises distribution is a normal distribution on a 
circle~\cite{weissteinvonMises}. The un-normalized von Mises distribution, 
$K_{\theta + \frac{\pi}{2}}(x,y)$ is defined as~\cite{Russell_etal14},

\begin{equation}
K_{\theta + \frac{\pi}{2}}(x,y) = \frac{\exp\big[ (\sqrt{x^2 + y^2} - R_{0})\sin({\tan^{-1}(\frac{y}{x})-(\theta + \frac{\pi}{2}}) \big]}{I_{0}(\sqrt{x^2 + y^2} - R_{0})}
\label{eq:FGO-von-mises}
\end{equation}
where $R_{0} = 2$ pixels is the radius of the circle on which the von Mises 
distribution is defined, $\theta + \frac{\pi}{2}$ is the angle at which the 
normal distribution is concentrated~\cite{weissteinvonMises} on the circle (also 
called mean direction), and $I_{0}$ is the modified Bessel function of the first 
kind. The distribution is then normalized as, 

\begin{equation}
\hat{K}_{\theta + \frac{\pi}{2}}(x,y) = \\ 
\frac{K_{\theta + \frac{\pi}{2}}(x,y)}{\max{(K_{\theta + \frac{\pi}{2}}(x,y))}}
\label{eq:FGO-von-mises-norm}
\end{equation}
$\hat{K}_{\theta - \frac{\pi}{2}}(x,y)$ is computed analogously. 

The BO pyramid, $B_{\theta + \frac{\pi}{2}, L}^{k}(x,y)$ for light objects on 
dark background capturing the BO activity for $\theta + \frac{\pi}{2}$ direction 
is computed as,
\begin{equation}
\begin{split}
\mathcal{B}_{\theta + \frac{\pi}{2} ,L}^{k}(x,y) = \text{max}\Bigg( 0, \mathcal{C}_{\theta}^{k}(x,y) \times \bigg(1 + \bigoplus_{j \geq k} \frac{1}{2^j} \hat{K}_{\theta + \frac{\pi}{2}}(x,y) \ast \mathcal{N}(\mathcal{CS}_{L}^{j}(x,y)) 
\\ - w_{opp} \bigoplus_{j \geq k} \frac{1}{2^j} \hat{K}_{\theta - \frac{\pi}{2}}(x,y) \ast \mathcal{N}(\mathcal{CS}_{D}^{j}(x,y)) \bigg)\Bigg) 
\label{eq:FGO-BOPyr-L}
\end{split}
\end{equation}

Similarly, the BO pyramid for $\theta + \frac{\pi}{2}$ direction for a dark 
object on light background is obtained by correlating normalized $\mathcal{CS}$ 
maps with $\hat{K}_{\theta \pm \frac{\pi}{2}}$ and summing the responses for all 
scales greater than the scale, $k$ at which BO map is being computed as,

\begin{equation}
\begin{split}
\mathcal{B}_{\theta + \frac{\pi}{2} ,D}^{k}(x,y) = \text{max}\Bigg( 0, \mathcal{C}_{\theta}^{k}(x,y) \times  \bigg(1 + \bigoplus_{j \geq k} \frac{1}{2^j} \hat{K}_{\theta + \frac{\pi}{2}}(x,y) \ast \mathcal{N}(\mathcal{CS}_{D}^{j}(x,y)) 
\\ - w_{opp} \bigoplus_{j \geq k} \frac{1}{2^j} \hat{K}_{\theta - \frac{\pi}{2}}(x,y) \ast \mathcal{N}(\mathcal{CS}_{L}^{j}(x,y)) \bigg)\Bigg)
\label{eq:FGO-BOPyr-D}
\end{split}
\end{equation} 
where, $w_{opp}$ is the synaptic weight for the inhibitory signal from the 
$\mathcal{CS}$ feature map of opposite contrast polarity. The symbol, 
$\bigoplus$ is used to denote pixel-wise addition of responses from all scales 
greater than $k$, by first up-sampling the response to the scale at which 
$\mathcal{B}_{\theta + \frac{\pi}{2} ,D}^{k}(x,y)$ is being computed. The other 
two pyramids, $\mathcal{B}_{\theta - \frac{\pi}{2} ,L}^{k}(x,y)$ and 
$\mathcal{B}_{\theta - \frac{\pi}{2} ,D}^{k}(x,y)$ for the opposite BO direction 
are computed analogously. 

With the BO pyramids related to dark and light $\mathcal{CS}$ pyramids already 
computed, we turn our attention to the computation of the local cue related BO 
pyramids. The local cue pyramids at different scales, $\mathcal{SA}_{\theta \pm 
\frac{\pi}{2}}^{k}(x,y)$ and $\mathcal{TJ}_{\theta \pm \frac{\pi}{2}}^{k}(x,y)$ 
are constructed, as explained in Sections~\ref{subsec:SAcomputation} and 
\ref{subsec:TJcomputation}, by successively down-sampling the local cue maps 
computed at native image resolution. Both local cues excite $\mathcal{B}$ cells 
of the same BO direction and inhibit the opposite BO direction $\mathcal{B}$ 
cells. 

The BO pyramid for $\theta + \frac{\pi}{2}$ BO direction related to the local 
cue, SA denoted as, $\mathcal{B}_{\theta + \frac{\pi}{2} ,SA}^{k}(x,y)$ is 
computed as,

\begin{equation}
\begin{split}
\mathcal{B}_{\theta + \frac{\pi}{2},SA}^{k}(x,y) = \text{max}\Bigg(0, \mathcal{C}_{\theta}^{k}(x,y) \times \bigg(1 + \displaystyle\bigoplus_{j \geq k} \frac{1}{2^j} \hat{K}_{\theta + \frac{\pi}{2}}(x,y) \ast \mathcal{N}(\mathcal{SA}_{\theta + \frac{\pi}{2}}^{j}(x,y)) 
\\ - w_{opp} \bigoplus_{j \geq k} \frac{1}{2^j} \hat{K}_{\theta - \frac{\pi}{2}}(x,y) \ast \mathcal{N}(\mathcal{SA}_{\theta - \frac{\pi}{2}}^{j}(x,y))\bigg)\Bigg)
\label{eq:FGO-BOPyr-SA}
\end{split}
\end{equation}
where we can see the SA cell ($\mathcal{SA}_{\theta + \frac{\pi}{2}}^{k}(x,y)$) 
having same BO preference as the BO cell, $\mathcal{B}_{\theta + \frac{\pi}{2} 
,SA}^{k}(x,y)$ has an excitatory effect on the BO cell, but 
$\mathcal{SA}_{\theta - \frac{\pi}{2}}^{k}(x,y)$ has an inhibitory effect. The 
synaptic weight, $w_{opp}$ remains unchanged as in Eqs~\ref{eq:FGO-BOPyr-L} and 
\ref{eq:FGO-BOPyr-D}. The BO pyramid, $\mathcal{B}_{\theta - \frac{\pi}{2}, 
SA}^{k}(x,y)$ related to SA, for opposite BO direction is computed in the same 
way. 

The BO pyramid related to T-Junctions for the BO direction, $\theta + 
\frac{\pi}{2}$ is computed as,

\begin{equation}
\begin{split}
\mathcal{B}_{\theta + \frac{\pi}{2},TJ}^{k}(x,y) &= \\
& \text{max}\Bigg( 0, \mathcal{C}_{\theta}^{k}(x,y) \times \bigg(1 + \bigoplus_{j \geq k} \frac{1}{2^j} \hat{K}_{\theta + \frac{\pi}{2}}(x,y) \ast \mathcal{N}(\mathcal{TJ}_{\theta + \frac{\pi}{2}}^{j}(x,y)) 
\\ &- w_{opp} \displaystyle\bigoplus_{j \geq k} \frac{1}{2^j} \hat{K}_{\theta - \frac{\pi}{2}}(x,y) \ast \mathcal{N}(\mathcal{TJ}_{\theta - \frac{\pi}{2}}^{j}(x,y)) \bigg)\Bigg)
\label{eq:FGO-BOPyr-TJ}
\end{split}
\end{equation}
The corresponding T-Junction pyramid for the opposite BO direction, $\theta - 
\frac{\pi}{2}$, denoted as $\mathcal{B}_{\theta - \frac{\pi}{2},TJ}^{k}(x,y)$ is 
computed analogously. 

The combined BO pyramid for direction, $\theta + \frac{\pi}{2}$ is computed by 
summing global and local cue specific BO pyramids as,

\begin{equation}
\begin{split}
\mathcal{B}_{\theta + \frac{\pi}{2}}^{k}(x,y) &= \alpha_{ref}\bigg(\mathcal{B}_{\theta + \frac{\pi}{2} ,L}^{k}(x,y) + \mathcal{B}_{\theta + \frac{\pi}{2} ,D}^{k}(x,y)\bigg) \\
 &+ \alpha_{SA} \bigg( \mathcal{B}_{\theta + \frac{\pi}{2} ,SA}^{k}(x,y) \bigg) + \alpha_{TJ} \bigg( \mathcal{B}_{\theta + \frac{\pi}{2} ,TJ}^{k}(x,y) \bigg)
\label{eq:FGO-BOPyr-SA-TJ}
\end{split}
\end{equation}
where $\alpha_{ref}$, $\alpha_{SA}$ and $\alpha_{TJ}$ are weights such that 
$\alpha_{ref} + \alpha_{SA} + \alpha_{TJ} = 1$, that control the contribution of 
$\mathcal{CS}$, $\mathcal{SA}$ and $\mathcal{TJ}$ cues to the BO response at 
that location respectively. By setting the weights to 0 or 1, we can study the 
effect of individual cue on BO response. It should be noted that the local cues 
are active only for the Orientation channel, so for the other channels, 
$\alpha_{SA}$ and $\alpha_{TJ}$ will be set to zero, by default. In the absence 
of local cues, combination of light and dark BO pyramids (first term in 
Eq~\ref{eq:FGO-BOPyr-SA-TJ}) results in contrast polarity invariant BO response. 
The corresponding BO pyramid for opposite BO preference, $\mathcal{B}_{\theta - 
\frac{\pi}{2}}^{k}(x,y)$ is computed as in Eq~\ref{eq:FGO-BOPyr-SA-TJ} by 
summing the light, dark and local cue BO pyramids of opposite BO preference.

Since the BO responses, $\mathcal{B}_{\theta \pm \frac{\pi}{2}}^{k}(x,y)$, are 
computed for each orientation, $\theta$ there will be multiple BO responses 
active at a given pixel location. But the boundary between figure and ground can 
only belong to the figure side, \ie there can only be one winning BO response 
for a given location. So, the winning BO response, denoted as 
$\widehat{\mathcal{B}}^{k}_{\theta + \frac{\pi}{2}}(x,y)$ is computed as,

\begin{equation}
\widehat{\mathcal{B}}^{k}_{\theta + \frac{\pi}{2}}(x,y) =
    \begin{cases}
      \max\bigg(0,\mathcal{B}^{k}_{\theta + \frac{\pi}{2}}(x,y) - \mathcal{B}_{\theta - \frac{\pi}{2}}^{k}(x,y)\bigg), & \text{if } \theta = \widehat{\theta} \\
      0, & \text{otherwise}
    \end{cases}
\label{eq:FGO-Winning-BO-Pyr}
\end{equation}
where $\widehat{\theta} = arg \max_{\theta}(\left|\mathcal{B}^{k}_{\theta + 
\frac{\pi}{2}}(x,y) - \mathcal{B}_{\theta - \frac{\pi}{2}}^{k}(x,y)\right|)$ is 
the orientation for which absolute difference between antagonistic pair of BO 
responses is maximum over all orientations. This gives the edge orientation at 
that location. So, the winning BO pyramid, $\widehat{\mathcal{B}}^{k}_{\theta + 
\frac{\pi}{2}}$ has non-zero response at a location only if the difference 
between the corresponding pair of BO responses for $\widehat{\theta}$ is 
non-negative. The winning BO pyramid, $\widehat{\mathcal{B}}^{k}_{\theta - 
\frac{\pi}{2}}$ for the opposite direction is computed analogously. 

Upto this point, the computation for all feature channels is identical. Now, if 
we denote the feature specific winning BO pyramid for $\theta + \frac{\pi}{2}$ 
direction for the Color channel by ${}_{C}\widehat{\mathcal{B}}^{k}_{\theta + 
\frac{\pi}{2}}$, Intensity feature channel by 
${}_{I}\widehat{\mathcal{B}}^{k}_{\theta + \frac{\pi}{2}}$ and Orientation 
feature channel by ${}_{O}\widehat{\mathcal{B}}^{k}_{\theta + \frac{\pi}{2}}$, 
then the final BO map, $\widetilde{\mathcal{B}}_{\theta + \frac{\pi}{2}}(x,y)$ 
for $\theta + \frac{\pi}{2}$ BO direction is computed by linearly combining the 
up-sampled feature specific BO maps across scales as,

\begin{equation}
\widetilde{\mathcal{B}}_{\theta + \frac{\pi}{2}}(x,y) = \bigoplus_{k=1}^{N_{s}} \bigg({}_{C}\widehat{\mathcal{B}}^{k}_{\theta + \frac{\pi}{2}}(x,y) + {}_{I}\widehat{\mathcal{B}}^{k}_{\theta + \frac{\pi}{2}}(x,y) + {}_{O}\widehat{\mathcal{B}}^{k}_{\theta + \frac{\pi}{2}}(x,y) \bigg) 
\label{eq:FGO-Final-BO-maps}
\end{equation}
where $\bigoplus$ represents pixel-wise addition of feature specific BO 
responses across scales after up-sampling each map to native resolution of the 
image. Similarly, $\widetilde{\mathcal{B}}_{\theta - \frac{\pi}{2}}$ is computed 
for $\theta - \frac{\pi}{2}$ BO direction. As we can see in 
Eq~\ref{eq:FGO-Final-BO-maps}, the contribution of every feature channel to the 
final BO map is the same, \ie, feature combination is equally weighted. Ten 
spatial scales ($N_{s} = 10$) are used. All parameters of the model are 
summarized in Table~\ref{tab:Modelparams}. In the end, we get 16 BO maps at 
image resolution, 8 each for $\theta + \frac{\pi}{2}$ and $\theta - 
\frac{\pi}{2}$ BO directions respectively.

\section{Adding local cues}
\label{sec:localCueComp}
Both local cues, SA and T-Junctions are computed at the native resolution of 
the images, but they influence BO cells of all scales as described in 
Eqs.~\ref{eq:FGO-BOPyr-SA}, \ref{eq:FGO-BOPyr-TJ}. In other words, the cues are 
computed once based on the analysis local image neighborhood, 
but their effect is not local\footnote{Should the effect of local cues also 
be local? See Section~\ref{sec:CraftNatura-Discussion} for related discussion}. 

\subsection{Computation of Spectral Anisotropy}
\label{subsec:SAcomputation}
\begin{figure}
	\centering
	\includegraphics[width=1.0\textwidth]{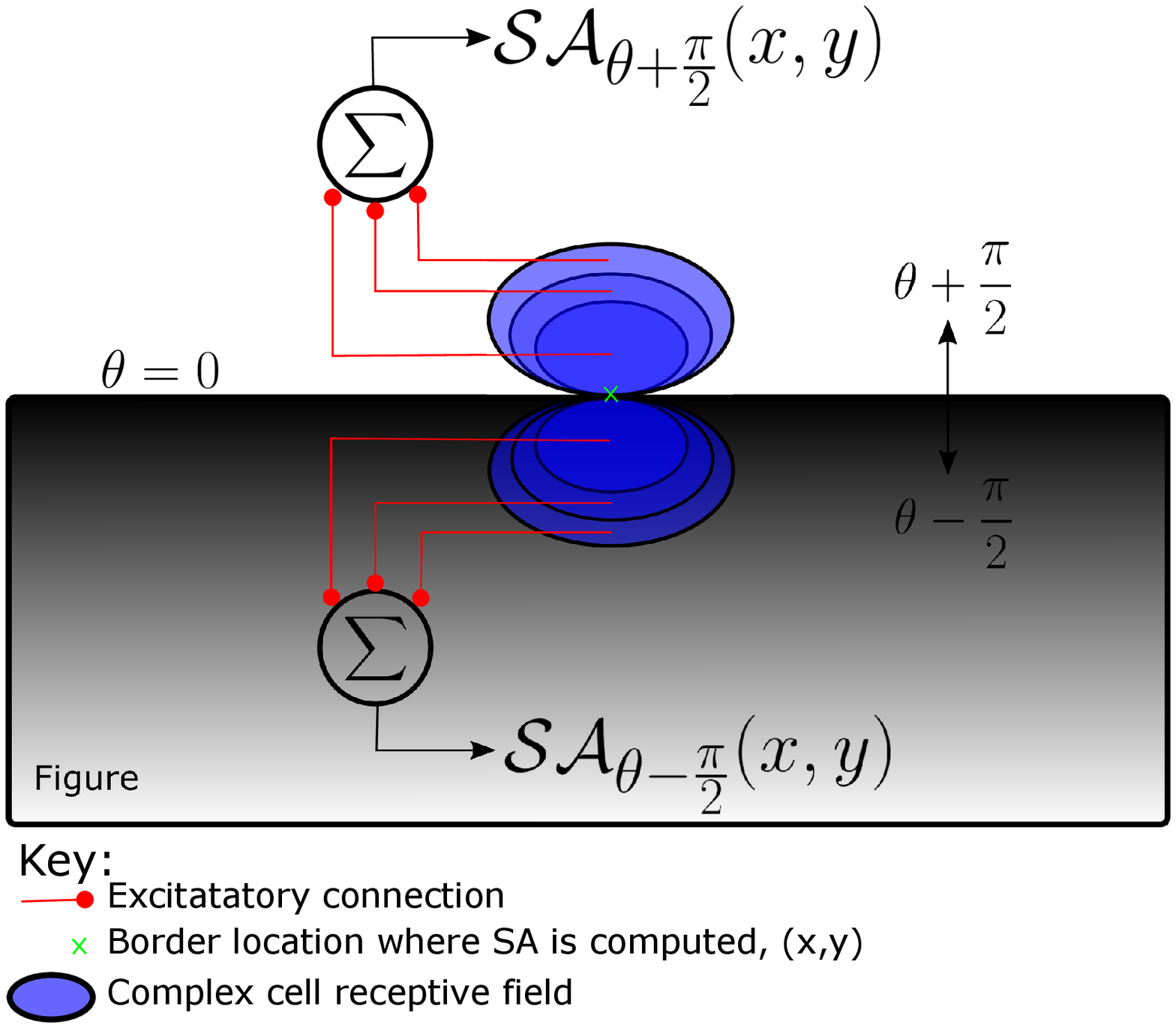}
	\caption{Biologically plausible computation of Spectral Anisotropy by pooling Complex 
cell responses. The local orientation, $\theta$ of the border between figure and 
ground at boundary location $(x,y)$ is 0. SA is computed at $(x,y)$ for two two 
opposite BO directions, $\theta + \frac{\pi}{2}$ and $\theta - \frac{\pi}{2}$. 
There is a  vertical intensity gradient on the figure side along the horizontal 
edge. By pooling the complex cell responses at various scales (hence, different 
spatial frequencies) on the side of an edge, we can quantify the intensity and 
texture gradients in the direction orthogonal to the edge orientation
}
	\label{fig:biolSA-Illus}
\end{figure}

Spectral Anisotropy, a local cue for FGO, that captures intensity and texture 
gradients very close to object boundaries, is computed by pooling Complex cell 
responses of various spatial frequencies from small image regions on either 
sides of the boundary (Figure~\ref{fig:biolSA-Illus}). This computation is neurally/biologically plausible.

SA at any location, $(x,y)$ in the image, for a specific orientation, $\theta$ 
and for the BO direction, $\theta + \frac{\pi}{2}$ is computed for one side of 
the border (side determined by the BO direction, $\theta + \frac{\pi}{2}$) as,

\begin{equation}
\label{eq:SACompPlus}
\mathcal{SA}_{\theta + \frac{\pi}{2}} (x,y) = \sum_{\omega_{r}} \mathcal{C}_{\theta}(x_{r_+}, y_{r_+},\omega_{r}) 
\end{equation}
where $\omega_{r} = \frac{\pi \times n_{lobes}}{2r}$. The Complex cell response, 
$\mathcal{C}_{\theta}(x_{r_+}, y_{r_+},\omega_{r})$, is computed as explained in 
Section~\ref{subsubsec:FGO-feat-comp-vis-O}, but with a different set of 
parameters, $\sigma_{SA}$, $\gamma_{SA}$, $\omega_{r}$ instead of $\sigma$, 
$\gamma$ and $\omega$ respectively. The values of $\sigma_{SA}$, $\gamma_{SA}$, 
$\omega_{r}$ and other relevant parameters are listed in 
Table~\ref{tab:SAparams}. Filter size is equal to $2r$ and $r$ is the 
perpendicular distance between the point, $(x,y)$ at which SA is being computed 
and the center of the Gabor filters. The centers of even and odd symmetric Gabor 
filters, hence the Complex cell are all located at  $(x_{r_+}, y_{r_+})$, from 
where the complex cell responses are pooled to compute SA. The term, $n_{lobes}$ 
determines the number of lobes in the Gabor filters. It is 2 or 4 for even 
symmetric Simple cells and 3 or 5 for odd symmetric Simple cells. The location 
from which Complex cell responses are pooled, $(x_{r_+}, y_{r_+})$  is computed 
as,

\begin{equation}
\label{eq:SACompXrPlus}
x_{r_+} = x + r \cos(\theta + \frac{\pi}{2})
\end{equation}

\begin{equation}
\label{eq:SACompYrPlus}
y_{r_+} = y + r \sin(\theta + \frac{\pi}{2})
\end{equation}

Similarly, SA at the same location, $(x,y)$, but for the opposite side of border 
at the same orientation, $\theta$ is computed as,
\begin{equation}
\label{eq:SACompMinus}
\mathcal{SA}_{\theta - \frac{\pi}{2}} (x,y) = \sum_{\omega_{r}} \mathcal{C}_{\theta}(x_{r_-}, y_{r_-},\omega_{r}) 
\end{equation}
where,
\begin{equation}
\label{eq:SACompXrMinus}
x_{r_-} = x + r \cos(\theta - \frac{\pi}{2})
\end{equation}

\begin{equation}
\label{eq:SACompYrMinus}
y_{r_-} = y + r \sin(\theta - \frac{\pi}{2})
\end{equation}

So, for every location there will be two SA cells capturing the spatial 
intensity and texture gradients on the two sides abutting the border. It has to 
be noted that the major axis orientation of the Gabor filters is the same as the 
local border orientation, $\theta$. This is because we want to capture the 
variation of spectral power in a direction orthogonal to the object boundary, 
which is captured by the Complex cells with their orientation parallel to the 
object boundary. This biologically plausible computation of SA with Complex 
cells responses captures the anisotropic distribution of high frequency spectral 
power on figure side we observed in \cite{ramenahalli2014local}. The SA maps 
thus obtained are decomposed into multiscale pyramids, $\mathcal{SA}_{\theta \pm 
\frac{\pi}{2}}^{k}(x,y)$, where superscript, $k$ denotes scale, by successive 
downsampling, which are used to compute the cue specific BO pyramids as 
explained in Section~\ref{subsec:FGO-multi-res-BO-pyr}, 
Equation~\ref{eq:FGO-BOPyr-SA}.

\begin{table}
\centering
\begin{tabular}{|l|l|}
\hline
\bf{Parameter} & \bf{Value} \\ \hline \hline
Min Filter Size & 9 \\ \hline
Max Filter Size & 25 \\ \hline
Filter Size Increment Step & 2 \\ \hline
Aspect Ratio ($\gamma_{SA}$) & 0.8 \\ \hline
$n_{lobes}$ (Simple Even cells, $S_e$) & 4 \\ \hline
$n_{lobes}$ (Simple Odd cells, $S_o$) & 5 \\ \hline
Std dev (Gaussian) ($\sigma_{SA}$) & $0.6 \times r$ \\ \hline
\end{tabular}
\caption{Parameters related to the Simple (Eqs~\ref{eq:SimpleEven} and 
\ref{eq:SimpleOdd}) and Complex (Eq~\ref{eq:ComplexCellResp}) cells used in 
Spectral Anisotropy computation
}
\label{tab:SAparams}
\end{table}

\subsection{Detecting of T-Junctions}
\label{subsec:TJcomputation}

\begin{figure}
	\centering
	\includegraphics[width=1.0\textwidth]{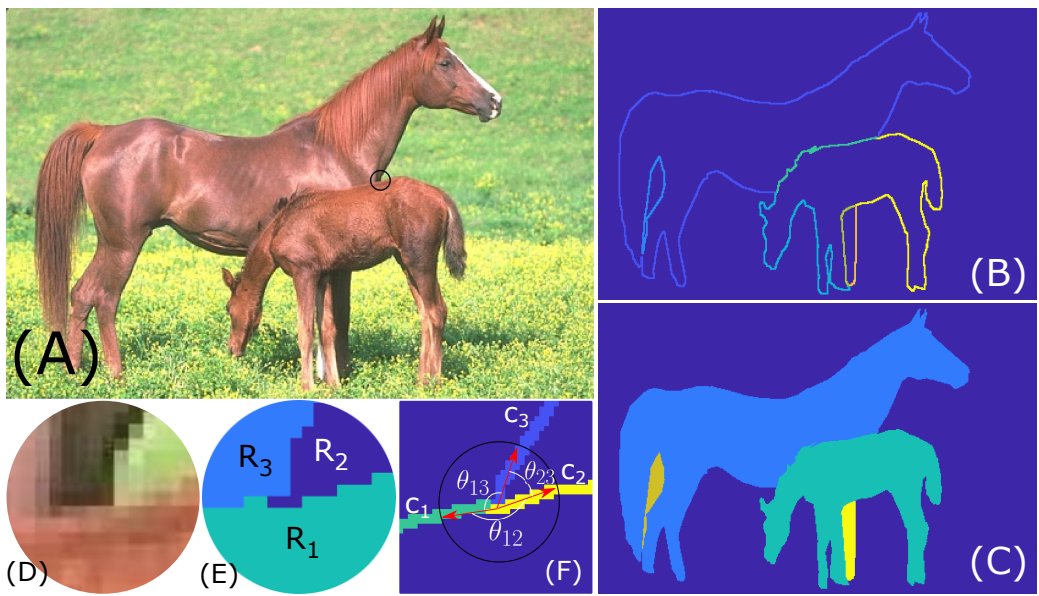}
	\caption{T-Junctions: Image (A) with T-Junction (black circle), the corresponding 
contours (B) and segments (C) are shown. Area based T-Junction determination: In 
(D), a small patch from image used for determining T-Junctions is shown. (E) 
Areas of three regions, $R_1$, $R_2$ and $R_3$ meeting at the T-Junction are 
determined. Contours abutting the segment ($R_1$) with largest area form the 
``hat'' of T-Junction. Angle based T-Junction determination: (F) From the 
junction location, 7 pixels are tracked for each contour, $c_1$, $c_2$ and 
$c_3$. Three vectors (red arrows) are defined based on the start (always 
junction location) and end points for each contour. The angles between the three 
vectors are determined. Contours for which largest angle ($\theta_{12}$) is 
observed form the ``hat'' of the T-Junction. Only matching T-Junctions based on 
segment area and contour angle are used in the model
}
	\label{fig:TJCompIllus2}
\end{figure}

The object edges and the regions bound by those edges called 
``segments'' are obtained using the gPb+ucm+OWT image segmentation 
algorithm~\cite{amfm_pami2011}, referred to as the gPb algorithm in other parts 
of this work. Image segmentation, partitioning of an image into disjoint 
regions, is considered as a pre-processing step occurring prior to FGO. The 
edges obtained using the gPb algorithm are represented as a Contour 
Map as shown in Figure~\ref{fig:TJCompIllus2}~(B). The corresponding Segmentation Map is 
shown in Figure~\ref{fig:TJCompIllus2}~(C). The 
Contour Map has uniquely numbered pieces of contours that 
appear to meet at a junction location. The Segmentation 
Map contains uniquely numbered disjoint regions bound by the contours. The 
Contour Map and Segmentation Maps are just a convenient way of representing the edge information. Only the 
locations at which exactly 3 distinct contours meet in the Contour Map 
(Figure~\ref{fig:TJCompIllus2}~(B)) and correspondingly the locations at which 
exactly 3 distinct segments meet in the Segmentation Map 
(Figure~\ref{fig:TJCompIllus2}~(C)) are considered for T-Junction determination. 
Such locations can be easily determined from the Segmentation and Contour maps.

As shown in Figure~\ref{fig:TJCompIllus2}E and 3F, at each junction location we 
have three regions, $R_1$, $R_2$ and $R_3$ and contours, $c_1$, $c_2$ and $c_3$ 
meeting. At each such junction, a circular mask of $N_{mask}$ pixels is applied 
and the corresponding small patches of the segmentation map and contour map are 
used for further analysis. We determine the contours forming the ``hat'' of the 
T-Junction (foreground) and the corresponding figure direction in two different 
ways: (1) based on the area of regions meeting at junction location within the 
small circular disk around junction; (2) based on the angle between contours 
meeting at the junction location. Finally, only those junctions locations for 
which figure direction, as determined based on both methods, is matching are 
introduced into the FGO model as T-Junction local cues. Matching based on two 
different methods improves the overall accuracy in correctly identifying the 
``hat'' (foreground) and ``stem'' (background) of T-Junctions, in effect the 
correct figure direction.

The local neighborhood of T-Junction influence is set to be a circular region of 
radius 15 pixels. All the border pixels near the junction location within a 
radius of 15 pixels that belong to the ``hat'' of the T-Junction are set to +1 
for the appropriate BO direction. Remember that for each orientation, $\theta$ 
we will have two T-Junction maps, one for the BO preference direction, $\theta + 
\frac{\pi}{2}$ denoted as $\mathcal{TJ}_{\theta + \frac{\pi}{2}}$ and the other 
for the opposite BO preference, $\theta - \frac{\pi}{2}$ denoted as 
$\mathcal{TJ}_{\theta - \frac{\pi}{2}}$. A pixel in $\mathcal{TJ}_{\theta + 
\frac{\pi}{2}}(x,y)$ is set to +1 if the direction of figure, as determined by 
both methods (Sections~\ref{subsubsec:TJcompAreaBased} and 
\ref{subsubsec:TJcompAreaBased})  is $\theta + \frac{\pi}{2}$, \ie ``stem'' of 
the T-Junction is in the $\theta - \frac{\pi}{2}$ direction. Similarly, 
$\mathcal{TJ}_{\theta - \frac{\pi}{2}}(x,y)$ computed. The T-Junction maps thus 
obtained are decomposed into multiscale pyramids, $\mathcal{TJ}_{\theta \pm 
\frac{\pi}{2}}^{k}(x,y)$, where superscript, $k$ denotes scale, by successive 
downsampling, which are used to compute the cue specific BO pyramids as 
explained in Section~\ref{subsec:FGO-multi-res-BO-pyr}, 
Equation~\ref{eq:FGO-BOPyr-TJ}.

\subsubsection{Area based T-Junction determination}
\label{subsubsec:TJcompAreaBased}

Let $R_1$, $R_2$ and $R_3$ be the three regions at a junction location $(x,y)$ 
(Figure~\ref{fig:TJCompIllus2}E). After extracting the circular region around 
the junction by applying a circular mask of radius, 6 pixels, we count the 
number of pixels belonging to each of the regions, $R_i$. The region, $R_i$ 
having the largest pixel count is determined as the figure region. In 
Figure~\ref{fig:TJCompIllus2}E, $R_1$ is the region with largest pixel count, 
hence determined as the foreground. The contours abutting the figure region, 
$R_1$ as determined by pixel count, which are $c_1$ and $c_2$ 
(Figure~\ref{fig:TJCompIllus2}F), form the ``hat'' of the T-Junction. Contour 
$c_3$ forms the ``stem'' of the T-Junction, which belongs to the background. 

The local orientation at each contour location is known. Vectors of length $1-3$ 
pixels, normal to the local orientation are drawn at each ``hat'' contour 
location within the $15 \times 15$ pixel neighborhood. If the normal vector 
intersects the figure region, $R_1$, as determined based on region area, the 
edge/contour location is given a value of +1 in the T-Junction map for the 
appropriate BO direction, which can be $\theta + \frac{\pi}{2}$ or $\theta - 
\frac{\pi}{2}$. This is done for every pixel in the edge/contour map within a 
neighborhood of 15 pixel radius around the T-Junction location for those 
contours ($c_1$ and $c_2$) that form the ``hat'' of the T-Junction. For example, 
in Figures~\ref{fig:TJCompIllus2}E and 3F, if the local orientation of $c_1$ and 
$c_2$ is roughly 0, then the end point of normal vector in the $\theta - 
\frac{\pi}{2}$ direction intersects with the figure region, $R_1$, as determined 
based on the segment area. So, the T-Junction map for $\theta - \frac{\pi}{2}$ 
BO preference direction is set to +1 within the circular neighborhood of $15 
\times 15$ pixels. The T-Junction map for $\theta + \frac{\pi}{2}$ BO direction 
will be zero. 

\subsubsection{Angle based T-Junction determination}
\label{subsubsec:TJcompAngleBased}
In this method, as in Section~\ref{subsubsec:TJcompAreaBased}, a small circular 
patch of radius, 7 pixels is extracted from the contour map around T-junction 
location. Pixels belonging to each contour, $c_i$ meeting at the junction are 
labeled with a distinct number, so for each contour, $c_i$ we track the first 7 
pixels starting from the junction location. \textcolor{black}{Since the starting 
point for each contour, $c_i$ is the same, the total angle at junction location 
is $360^{o}$}. For each contour, we define a vector 
(red arrows in Figure~\ref{fig:TJCompIllus2}F) from the junction location to the 
last tracked point on the contour. We then compute the angle between the vectors 
corresponding to contours.  The contours between which angle is the largest form 
the ``hat'' of the T-junction. For example, in Figure~\ref{fig:TJCompIllus2}F, 
$\theta_{12}$ is the angle between $c_1$ and $c_2$, which is also the largest of 
the three angles, $\theta_{12}$, $\theta_{32}$ and $\theta_{13}$. So, in the 
angle based T-junction computation also, $c_1$ and $c_2$ are determined to form 
the ``hat'' of the T-junction. The figure direction at every pixel of the 
``hat'' contours is determined as in Section~\ref{subsubsec:TJcompAreaBased}. 

Among all the potential T-Junctions determined using the angle based 
method, potential Y-Junctions and Arrow junctions are discarded based on the 
angle formed by the contours at junction location. If the largest angle is 
greater than $180^{0}$, such junctions are discarded. Since the largest angle 
greater than $180^{0}$ is typically seen in the case of Arrow-junctions, we do 
not include them in the computation. Arrow junctions appear in a scene when the 
corner of a 3D 
structure is seen from outside. In the same way, if each angle at a junction 
location is within $120^{0} \pm 10^{0}$, such junctions are discarded as those 
are most likely Y-Junctions. Y-Junctions appear in a scene when a 3D corner is 
viewed from inside the object, for example, corner of a room viewed from inside 
the room. Rest of the T-Junctions are included in our computation. Angle based 
filtering of potential Arrow or Y-Junctions was not considered in previous 
methods~\cite{palou2011occlusion, palou2013monocular}.

T-Junctions and their figure directions are determined using both Segment Area 
based and Contour Angle based methods and the T-Junctions are incorporated into 
the model only in those cases, where both methods give matching figure 
direction, which makes T-Junction determination more accurate. 

Accurately determining the figure side of a T-junction from a small neighborhood 
of 6-7 pixel radius is quite challenging because, within that small neighborhood we 
generally do not have any information indicative of figure/ground relations, 
other than contour angle and segment area. Even though key point detection is a 
well studied area, hence locating a T-Junction is not problematic, deciding 
which of the three regions is the foreground based on information from a small neighborhood is extremely challenging. So, when locally 
determining figure side of a T-junction, segment area and contour angle were 
found to be the most exploitable properties. 

\section{Data and methods}
\label{sec:CraftNatural-Dataset}
The figure-ground dataset,  a subset of BSDS 300 dataset, consists of 200 images 
of size $321\times481$ pixels, where each image has two ground truth 
figure-ground labels~\cite{ren2006figure} and corresponding boundary maps. For 
each image, the two sets of figure-ground labels labels are annotated by users 
other than those who outlined the boundary maps. The figure-ground boundary 
consists of figure side of the boundary marked by +1 and the ground side 
boundary by -1. 

The figure-ground classification accuracy (FGCA) for an image we report is the percentage of the  
total number of boundary pixels in the ground truth figure/ground label map for 
which a correct figure/ground classification decision is made by the model 
described in Section~\ref{sec:FGO-Model}. Even though the model computes BO 
response at every location where $\mathcal{C}_{\theta}$ cells are active, the BO 
responses are compared only at those locations for which ground truth 
figure/ground labels exist.

Whenever the two ground truth label maps differ for the same image, average of 
the FGCA for both ground truth label maps is reported. Since different 
figure-ground labelers interpret figure and ground sides differently depending 
on the context, such differences arise, as a result, the self-consistency 
between figure-ground labelings between the two sets of ground truth annotations 
is 88\%, which is the maximum achievable FGCA for the dataset. At each pixel, 
the direction of figure, as determined by the model can be correct or wrong. So, 
the average FGCA for the entire dataset, at chance is 50\%, assuming 
figure/ground relations at neighboring pixels are independent. This assumption 
is consistent with previously reported results~\cite{ren2006figure}, where same 
assumption was made.  The complete details of the figure-ground dataset can be 
found in  \cite{martin2001database,ren2006figure,Fowlkes2007}. 

The entire BSDS figure/ground dataset consisting of 200 images is randomly split 
into training set of 100 images and test set of 100 images. Parameters of the 
model are tuned for the training dataset and the optimal values of parameters 
found for the training set are used to evaluate the FGCA of the test set of 
images. The average FGCA that we report for the entire test set is the average 
of FGCAs of all 100 images in the test set. 

\section{Results}
\label{sec:FGO-Results}
To remind the readers, the model with only global cues of convexity, 
surroundedness and parallelism, without any local cues is referred to as the 
\textit{Reference model}. As explained in Section~\ref{sec:FGO-Model}, local cues, SA and 
T-Junctions are added to the Orientation feature channel of the reference model. 
As we have previously described in Section~\ref{subsec:FGO-multi-res-BO-pyr}, by 
setting $\alpha_{SA} = 0 $ and $\alpha_{TJ} = 0$ in Eq~\ref{eq:FGO-BOPyr-SA-TJ}, 
the model with local cues can be reduced to the reference model. Similarly, by 
switching the weights for each local cue to zero, the effect of the other local 
cue on FGO can be studied. As explained in 
Section~\ref{subsec:FGO-multi-res-BO-pyr}, the winning BO pyramids are 
up-sampled to image resolution and summed across scales and feature channels 
(Eq~\ref{eq:FGO-Final-BO-maps}) for each BO direction to get the response 
magnitude for that BO direction. The BO information derived this way is compared 
against the ground-truth from BSDS figure/ground dataset. 

First, we wanted to quantify the performance of the reference model, which is 
devoid of both local cues, in terms of FGCA. With $\alpha_{SA} = 0$ and 
$\alpha_{TJ} = 0$, the overall FGCA for 100 test images was 58.44\% (standard deviation = 0.1146). With only 
global cues, the 58.44\% FGCA we achieved is 16.88\% above chance level (50\%). 
Hence, we can conclude that the global Gestalt properties of convexity, 
surroundedness and parallelism, which the reference model embodies, are 
important properties that are useful in FGO. The parameters used in the 
reference model computation are listed in Table~\ref{tab:Modelparams}. Unless 
stated otherwise explicitly, those parameters in Table~\ref{tab:Modelparams} 
remain unchanged for the remaining set of results that we are going to discuss. 
Only the parameters specifically related to the addition of local cues are 
separately tuned and will be explicitly reported. 

\begin{table}
\centering
\begin{tabular}{|l|l|}
\hline
\textbf{Parameter}     & \textbf{Value}  \\ \hline \hline
$\gamma$        & 0.5    \\ 
$\sigma$        & 2.24   \\ 
$\omega$        & 1.57   \\ 
$\sigma_{in}$ & 0.90   \\ 
$\sigma_{out}$ & 2.70   \\ 
$R_{0}$      & 2.0    \\ 
$w_{opp}$    & 1.0    \\ 
$\sigma_{1}$ & 3.2    \\ 
$\gamma_{1}$ & 0.8    \\ 
$\omega_{1}$ & 0.7854 \\ 
$N_{s}$             & 10    \\ \hline
\end{tabular}
\caption{Parameters of the Reference FGO model without any local cues}
\label{tab:Modelparams}
\end{table}

Next, we wanted to study the effect of adding each local cue individually 
(Sections~\ref{subsec:AddSA} and \ref{subsec:AddTJResults}) and then the effect 
of both local cues together (Section~\ref{subsec:TJandSAResults}). 

\subsection{Effect of adding Spectral Anisotropy}
\label{subsec:AddSA}
As explained in Section~\ref{subsec:SAcomputation}, Spectral Anisotropy was 
computed at the native resolution of the image by pooling Complex cell responses 
at many scales for each orientation. For each orientation, $\theta$, two SA 
maps, $\mathcal{SA}_{\theta + \frac{\pi}{2}}$ and $\mathcal{SA}_{\theta - 
\frac{\pi}{2}}$ are created for respective antagonistic BO directions with 
respect to $\theta$. The SA maps are then decomposed into multiscale pyramids by 
successively downsampling. The SA pyramids are then incorporated into the model 
as explained in Eq.\ref{eq:FGO-BOPyr-SA} and Eq.\ref{eq:FGO-BOPyr-SA-TJ}. In 
this case, parameters $\alpha_{ref}$ and $\alpha_{SA}$ are tuned for the 
training dataset and $\alpha_{TJ}$ is set to 0. 

The parameter tuning procedure we use here is the same for other cases as well. 
We use multi-resolution grid search for parameter tuning with the condition that 
the sum of tuned parameters should be 1. In this case, the condition was 
$\alpha_{ref} + \alpha_{SA} = 1$. We stop refining the resolution of the grid 
when the variation in FGCA upto second decimal point is zero, \ie, only small 
changes are seen from third digit onward, after the decimal point. 

The optimal parameters were found to be, $\alpha_{ref} = 0.35$ and $\alpha_{SA} 
= 0.65$ for the training dataset. With these optimal parameter values, the FGCA 
for the test set was 62.69\% (std. dev = 0.1204), which is a 7.3\% improvement in the model's 
performance after adding the local cue, Spectral Anisotropy, compared to the 
reference model's FGCA of 58.44\%. To verify if the improvement in FGCA that we 
see is statistically significant, we performed an unpaired sample, right tailed 
t-test (Table~\ref{tab:testResultsSummary}), where the null hypothesis was that 
the means of FGCAs of the reference model and the model with SA are equal. The 
alternate hypothesis was that the mean FGCA of the model with SA is higher than 
that of the reference model. The significance level, $\alpha = 0.05$ was chosen. 
For other results (Sections~\ref{subsec:AddTJResults}, 
\ref{subsec:TJandSAResults}) as well, we do the same type of test, where the 
reference model's FGCA is compared with that of modified model's FGCA having 
different local cues. Hereafter, we refer to them as \textit{statistical tests}. 

Statistical tests show that the mean FGCA of the model with SA is significantly 
higher than that of the reference model ($p = 5.2 \times 10^{-301}$). This 
demonstrates SA is a useful cue and can be successfully incorporated into the 
reference model, adding which results in statistically significant improvement 
in the model's performance. This, and all other results are summarized in 
Table~\ref{tab:testResultsSummary} for the test dataset.

\subsection{Effect of adding T-Junctions}
\label{subsec:AddTJResults}
As described in Section~\ref{subsec:TJcomputation}, T-Junctions are computed at 
image resolution using the segmentation map and edge map obtained using the 
gPb~\cite{amfm_pami2011} algorithm. Each of the T-Junction maps for the 16 
different BO directions is successively downsampled to create multiscale 
T-Junction pyramids. The T-Junction pyramids are incorporated into the model as 
explained in Eq.\ref{eq:FGO-BOPyr-TJ} and Eq.\ref{eq:FGO-BOPyr-SA-TJ} and by 
setting $\alpha_{SA} = 0$. The other two parameters, $\alpha_{ref}$ and 
$\alpha_{TJ}$ are tuned on the training dataset. With optimal parameter values, 
$\alpha_{ref} = 0.03$ and $\alpha_{TJ} = 0.97$ (and $\alpha_{SA} = 0$), the FGCA 
for the test set was found to be 59.48\% (std. dev. = 0.1127). Compared to the reference model's FGCA 
of 58.44\%, we see that adding T-Junctions improves the model's performance in terms of FGCA by 1.78\%. Based 
on the statistical tests (Table~\ref{tab:testResultsSummary}), we find that the 
improvement in FGCA that we see is indeed statistically significant. 

\subsection{Effect of adding both Spectral Anisotropy and T-Junctions}
\label{subsec:TJandSAResults}
SA is computed as explained in 
Section~\ref{subsec:SAcomputation}, T-Junctions are computed as explained in 
Section~\ref{subsec:TJcomputation}, where T-Junctions are derived from 
automatically extracted edges using the gPb algorithm. Both cues are added to 
the Reference model according to Eq~\ref{eq:FGO-BOPyr-SA-TJ}. The parameters 
$\alpha_{ref}$, $\alpha_{SA}$ and $\alpha_{TJ}$ are tuned simultaneously on the 
training dataset using multiresolution grid search as before,  with the 
constraint, $\alpha_{ref} + \alpha_{SA} + \alpha_{TJ} = 1$. The optimal values 
of the parameters were found to be, $\alpha_{ref} = 0.05$, $\alpha_{SA} = 0.15$ 
and $\alpha_{TJ} = 0.80$. All other parameters remained unchanged as shown in 
Table~\ref{tab:Modelparams}. The FGCA of the combined model with both local 
cues, Spectral Anisotropy and T-Junctions was 63.57\% (std. dev = 0.1179)
for the test dataset, which is higher that the FGCAs we obtained for the 
individual cues when they were added separately. We see an improvement in FGCA 
of 8.78\% compared to that of the reference model with no local cues. As before, 
an unpaired sample, right tailed t-test comparing the reference model's 
figure/ground decisions and the combined model's figure/ground decisions with 
both SA and T-Junctions showed 
statistically significant improvement (Table~\ref{tab:testResultsSummary}). 

In addition to comparing the performance of the model with both 
local cues with 
the Reference model, we also compared the performance of the model with both 
local cues (Ref model + SA + T-Junctions) to the model with only one (Ref model 
+ SA) local cue. Unpaired sample right-tailed t-tests were used again with a 
significance level of 0.05. In this case the the null hypothesis is that adding 
T-Junctions to the Reference Model with SA does not lead to statistically 
significant improvement in FGCA. The alternate hypothesis is that adding 
T-Junctions leads to statistically significant improvement in FGCA when compared 
to the FGCA of Reference (global cues only) + SA model. Tests show adding 
T-Junctions to the Reference + SA model leads to a statistically significant 
improvement (p = $1.8911 \times 10^{-17}$).

\begin{table}
\centering
\begin{tabular}{|l|l|l|l|l|}
\hline
\multicolumn{1}{|l|}{} & \begin{tabular}[c]{@{}l@{}}\textbf{FGCA} \\ (std. dev)\end{tabular} & \begin{tabular}[c]{@{}l@{}}\textbf{\%age}\\  \textbf{increase}\end{tabular} & \begin{tabular}[c]{@{}l@{}}\textbf{Stat} \\ \textbf{Sig?}\end{tabular} & \textbf{p-value}  \\ \hline \hline
\textbf{Reference Model}  & \begin{tabular}[c]{@{}l@{}}58.44\% \\ (0.1146)\end{tabular}  & -   & -   & -  \\ \hline
\textbf{With SA}          & \begin{tabular}[c]{@{}l@{}}62.69\% \\ (0.1204)\end{tabular} & 7.3\%   & Yes       & $5.2 \times 10^{-301}$ \\ \hline
\begin{tabular}[c]{@{}l@{}}\textbf{With T-Junctions} \\ (gPb~\cite{amfm_pami2011} based boundaries)\end{tabular} & \begin{tabular}[c]{@{}l@{}}59.48\% \\ (0.1127)\end{tabular} & 1.78\% & Yes & $3.38 \times 10^{-26}$ \\ \hline
\begin{tabular}[c]{@{}l@{}}\textbf{With SA and T-Junctions} \\ (gPb~\cite{amfm_pami2011} based boundaries)\end{tabular} & \begin{tabular}[c]{@{}l@{}}63.57\% \\ (0.1179)\end{tabular} & 8.78\% & Yes & 0  \\ \hline
\end{tabular}
\caption[Summary of results for the FGO model with local and global cues for the 
BSDS test dataset]{Summary of results for the test dataset: Adding SA to the reference model 
improves the FGCA by 7.3\%. With T-Junctions derived from automatically 
extracted edges, the FGCA improvement is 1.78\%. Each individual local cue, added alone, produces 
statistically significant improvement in model performance, in terms of FGCA. 
When both are added together, the FGCA observed is higher than that we see with 
individual local cues, indicating the local cues are mutually facilitatory. Numbers within parentheses in Column~2 represent the standard deviation of FGCA. 
All results are statistically significant
}
\label{tab:testResultsSummary}
\end{table}

\begin{figure}
	\centering
	\includegraphics[width=1.0\textwidth]{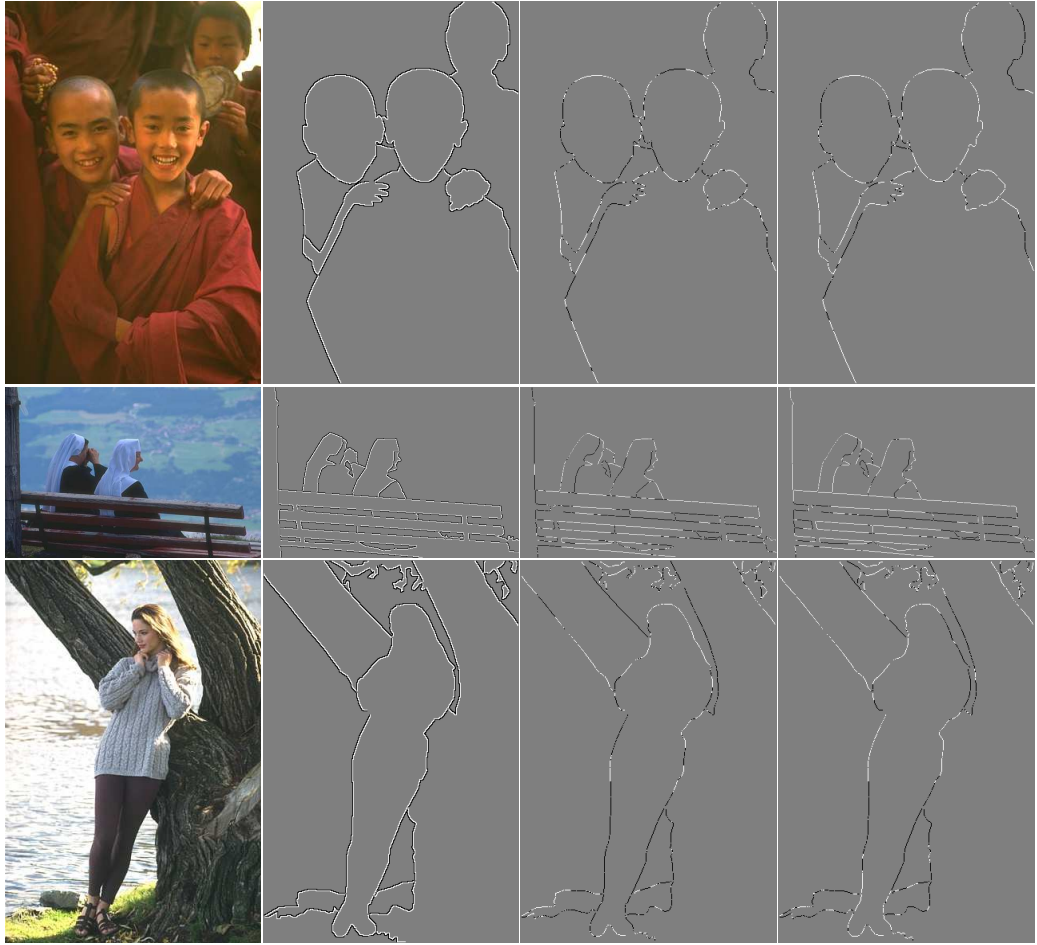}
	\caption[Illustration of Figure/Ground classification results in a few example 
	images]{Figure/Ground classification results in a few example images: For the images in 
the first column, the figure/ground ground-truth maps are shown in column 2, 
where a white pixel denotes the figure side of the border, black pixel, the 
ground side. Column~3 shows the figure/ground classification map for the 
reference model with no local cues. Column 4 images represent figure/ground 
classification maps for the model with both local cues, Spectral Anisotropy and 
T-Junctions, where T-Junctions are derived from automatically extracted edges. In images of columns 3--4, if a white pixel 
on the gray background indicates that a correct figure/ground decision was made 
by the model at that location, a black pixel indicates it was wrong, in 
comparison to the ground truth. 
}
	\label{fig:ResultsRefSATJ_2}
\end{figure}

\begin{figure}
	\centering
	\includegraphics[width=1.0\textwidth]{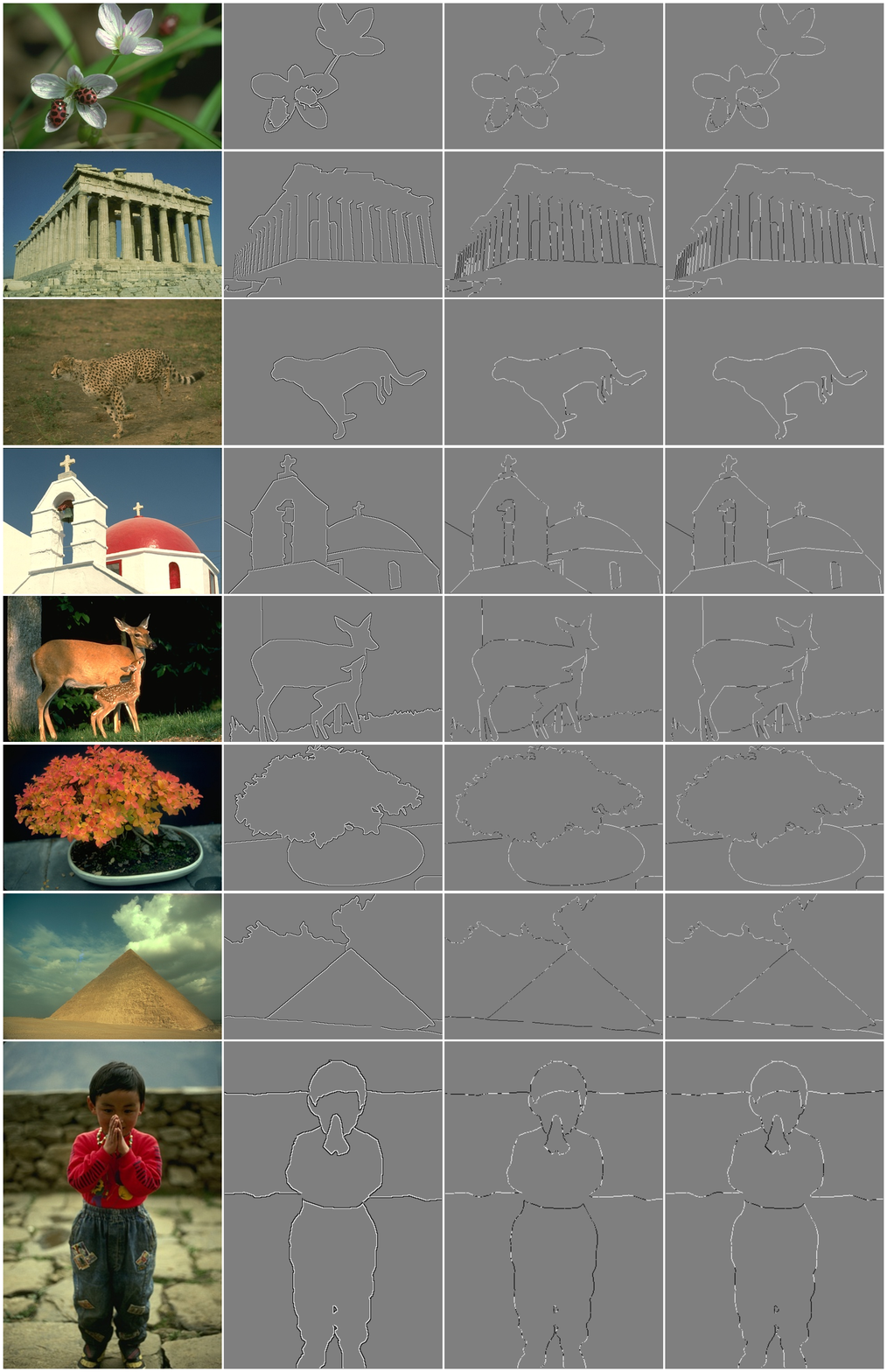}
	\caption[A few more examples of figure/ground classification results]{A few more 
	examples of figure/ground classification results. The different 
columns of images here are arranged in the same order as in 
Figure~\ref{fig:ResultsRefSATJ_2}}
	\label{fig:ResultsRefSATJ_1}
\end{figure}

In summary, we show that both SA and T-Junctions are useful local cues of FGO, 
which produce statistically significant improvement in FGCA when added alone. 
When both cues are simultaneously present, they lead to even higher improvement 
in FGCA of the model indicating the cues are mutually facilitatory. An 
improvement of $\approx 9\%$ with only a few local and global cues at a minimal 
computational cost (see \ref{subsec:TJCompComplexity} for computational cost 
analysis) is truly impressive. Figures~\ref{fig:ResultsRefSATJ_2} and 
\ref{fig:ResultsRefSATJ_1} show FGO results for some example images from the 
test dataset when both SA and T-Junctions are added.

Next, we compare the performance of our model with state of the art methods 
for which all steps are fully automated (see Table~~\ref{tab:FGOModelComparison}). Here, we are comparing the FGCA of 
the model with both local cues 
with other methods that are not neurally inspired, instead are learning 
based and trained on thousands of images. Our model performs better than that of 
\citet{maire2010simultaneous} even when it has only a few local and global FGO 
cues and not specifically tuned for best performance. The performance of our 
model is competitive with the state of the art models given our constraints 
discussed above, but leaves room for improvement. The performance of the model 
can be substantially improved by making several simple modifications and adding 
more local and global cues as discussed in 
Section~\ref{sec:CraftNatural-FutureWork}.

\begin{table}
\centering
\scalebox{1.0}{
\begin{tabular}{|l|l|}
\hline
\textbf{Algorithm}                         & \textbf{FGCA}   \\ \hline \hline
M. Maire etal, ECCV, 2010~\cite{maire2010simultaneous}         & 62\%  \\  \hline
\textbf{Our method}                        & \textbf{63.6}\%     \\ \hline
X. Ren etal, ECCV, 2006~\cite{ren2006figure}           & 68.9\%  \\ \hline
P. Salembier etal, IEEE TIP, 2013~\cite{palou2013monocular} & 71.3\%   \\ \hline
CL. Teo, etal, CVPR 2015~\cite{teo2015fast}          & 74.7\%               \\ \hline
D. Hoiem etal, ICCV 2007~\cite{hoiem2011recovering}          & 79\%           \\ \hline
\end{tabular}
}
\caption[FGO Model comparison with existing literature]{Comparison of FGCA of 
our model with existing fully automated FGO models: Our 
model performs better than \citet{maire2010simultaneous}, which uses 64 
different \textit{shapeme} based cues. \citet{ren2006figure} use empirically 
measure junction frequencies of 4 different junction types along with 
\textit{shapeme} cues in a CRF model. They compare with FG ground-truth on 
a partial set of edges only. Other models use a higher number of cues for FGO. With 
only a few local and global Gestalt cues, our neurally motivated, fully feed-forward 
model built with the purpose of studying effect of local cues, hence not optimized for best FGCA still performs 
competitively with existing models. As discussed in 
Chapter~\ref{sec:CraftNatural-FutureWork}, the model's FGCA can be substantially improved 
with some minimal modifications.
} 
\label{tab:FGOModelComparison}
\end{table}

\section{Discussion}
\label{sec:CraftNatura-Discussion}

We see $\approx 9\%$ improvement in FGCA of the combined model with both local cues. 
This improvement, from only two local cues added to one of the three feature 
channels is truly impressive. Moreover, the three feature channels (Color, Intensity and Orientation) were 
weighted equally. Better results can be achieved if we tune the 
weights for individual feature channels. But, since our objective here was to study how to 
integrate local and global cues and measure the relative importance of local cues in FGO, feature 
specific weight tuning was not done, but we consider to do this in future (See Section~\ref{sec:CraftNatural-FutureWork} for future work). 
Moreover, it is important to note that FGCA of the model with both local cues is 
always higher than the FGCAs of models with individual local cues. This suggests 
the local cues are mutually facilitatory, which is further 
validated by the fact that we see statistically significant improvement in FGCA 
when T-Junctions are added as an additional cue to the Reference model having SA 
as one of the local cues. 

We introduce a few novel methods in our work. First, demonstrating that Spectral Anisotropy can be computed with Simple 
and Complex cells found in area V1 is a novel contribution. The significance of 
this computation is that it demonstrates SA can be computed in low level visual 
areas, even in the striate cortex and it does not require specialized cells to 
detect these shading/texture gradients. Only a specific arrangement of Complex 
cells of various spatial frequencies on each side of the border is sufficient. 
These cues, first mathematically shown to be useful by \citet{Huggins_etal01}, 
were psychophysically validated by \citet{Palmer_Ghose08}. We showed these 
patterns are abundantly found in natural images~\cite{Ramenahalli2011a} and can 
be efficiently computed using 1D FFTs~\cite{ramenahalli2014local}. Now, we show 
that these cues can be computed in a biologically plausible manner, using 
Complex cells found commonly in striate cortex. 

Next, in the detection of 
T-Junctions, we filter out Y-Junctions and Arrow junctions using the angle 
property of these junction types. Since Y-Junctions and Arrow junctions are not 
occlusion cues, ideally those should not be considered as T-Junctions, hence we 
device a method to remove such junctions. To the best of our knowledge, previous 
methods~\cite{palou2011occlusion, palou2012local, palou2013monocular} that use 
T-Junctions as FGO cues have not looked closely at this issue, which we consider 
novel in our approach. Also, we explicitly compute the local figure/ground 
relations at a T-Junction local based on local information, which is new. And, 
the way we organize the local cue computation such that the same 
computational routine can be used for incorporation of both cues into the model 
is noteworthy. With this, the implementation is made more efficient, allowing 
easy parallelization using Graphics Processing Units (GPU) and other 
hardware. Moreover, the combination of features and local cues is done at a late 
stage (Eq~\ref{eq:FGO-BOPyr-SA-TJ}), which allows independent and parallel 
computation of features and local cues, which again makes the model 
computationally more efficient, allowing parallelization. 

Even though we see $\approx2\%$ FGCA 
improvement when T-junctions are added, it is a relatively small, but 
statistically significant, improvement compared to that adding SA. Since 
T-Junctions are generally regarded as strong cues of occlusion, this small, 
statistically significant improvement may seem counter-intuitive. But it is 
important to note that T-Junctions are extremely sparse, can be computed only at 
a few locations where exactly 3 different regions partially occlude each other, whereas SA can be 
computed at every border location of an object. Given the sparsity of 
T-junctions, they can still be considered stronger FGO cues compared to SA. The presence of 
``inverted'' T-Junctions~\cite{palou2013monocular,palou2012local,palou2011occlusion}, could 
also be the reason for diminished effect of T-Junctions. From a computational cost 
perspective (\ref{subsec:TJCompComplexity}), even though the cost is $O(N_{mask}^2)$, given their sparsity 
(typically 3-10 T-Junctions per image), adding them as a local cue is justified. 

Even though it is commonly assumed that T-Junctions are unambiguous cues of 
occlusion, no systematic, data-driven analysis of the utility of T-Junctions as a 
classic Gestalt cue was available until now.  Moreover, there are few instances 
where researchers argue from the opposite perspective. \citet{tse1998amodal} 
argue that high level surface and volume analysis takes 
place first, and only after such an analysis, a T-Junction is interpreted to be 
an occlusion cue. As a result, we may not consciously notice the prevalence of ``inverted'' 
T-Junctions. 

The traditional view that T-Junctions are unambiguous cues of occlusion has also 
been challenged by psychophysics experiments of 
\citet{mcdermott2004psychophysics}, where they find that making occlusion 
decisions from a small aperture, typically a few pixels wide, in real images is 
hard for humans. Some studies also suggest junctions in general, hence 
T-Junctions, can be cues for image segmentation, but not for occlusion 
reasoning~\cite{van2011bayesian}. These previous works and our own results do 
not support the generally held view that T-Junctions are the most unambiguous 
occlusion cues. But, these cues are useful and produce statistically significant 
improvement in FGCA. This is an important contribution of our work.

While comparing the performance of our model with existing methods, as noted in 
Section~\ref{subsec:TJandSAResults}, we need to keep in mind some important differences. First, our 
model is not trained on image features, hence generalization to any other 
dataset does not require additional training. Second, our model is neurally 
inspired, built to provide a general framework for incorporating and studying 
local and global Gestalt cues, not specifically optimized for best accuracy. 
Moreover, we use only a handful of cues, yet perform better than some existing 
models (\citet{maire2010simultaneous} in Table~\ref{tab:FGOModelComparison}). 
While \citet{maire2010simultaneous} 
uses 64 different \emph{shapemes}, descriptors of local shape derived from 
object edges, \citet{ren2006figure} incorporates empirical frequencies of 4 
different junction types derived from training data, in addition to shapemes 
in a Conditional Random Field based model. Also, 
\citet{ren2006figure} compare figure/ground relations with the ground-truth only 
at a partial set of locations where their edge detection algorithm finds a 
matching edge with the ground-truth. It is not clear what percentage of edges 
match with the ground-truth. \citet{palou2013monocular} use 8 color based cues, 
in addition to T-Junctions and local contour convexity in their model. The other 
two (\cite{teo2015fast,hoiem2011recovering} ) models use a much larger number of 
cues to achieve FGO. Moreover, the models we are comparing with are neither 
strictly Gestalt cue based nor neurally motivated. To the best of our knowledge, 
there are no comparable neurally inspired, feed-forward, fully automated models that 
are tested on the BSDS figure-ground dataset. The model proposed by 
\citet{sakai2012consistent} is tested on BSDS FG database, but it requires human 
drawn contours. 

The method we use to report FGCA can be very different from the methods of other models listed in 
Table~\ref{tab:FGOModelComparison}. We report the average of all pixels for all 
100 test images, which can be considerably lower than computing the FGCA image 
by image and then averaging the FGCA of all images. It is not clear from other methods in 
Table~\ref{tab:FGOModelComparison}, how the FGCA numbers were reported. Moreover, the exact split of the dataset 
into train and test set also has an effect. For some splits, the FGCA can be 
higher. The methods reported in 
Table~\ref{tab:FGOModelComparison} may not have used the same test/train split as it is not 
reported in previous methods. So, instead of comparing with existing methods in 
terms of absolute FGCA, a more appropriate way to look at our results would be 
from the perspective of relative improvement after adding each cue. From this perspective, we 
do see statistically significant improvement with the addition of each local cue. Moreover, our 
motivation in this study was always to quantify the utility of local and global 
cues and build a general framework to incorporate and study the effect of multiple local and 
global cues.

Lastly, we investigate if the influence of local cues should be 
strictly local or global. In our model, even though the local cues, SA and 
T-Junctions, are computed based on the analysis of a strictly local neighborhood 
around the object boundary, they modulate the activity of $\mathcal{B}$ cells at 
all scales, \ie, their influence is global in nature. Should the influence of 
local cues be also local? To answer this question, we added local cues only at 
the top 2 layers of the model, tuned the optimal parameters, $\alpha_{ref}$, 
$\alpha_{SA}$ and $\alpha_{TJ}$ accordingly and recomputed FGCA. We found that 
with local cue influence at only the top two layers, the FGCA we obtained was 
lower than having them at all scales (See \ref{app:2layerLocalCueFGCA} 
for details). This confirms the influence of local cues should not be local, even 
though their computation should be strictly local to reduce the computational cost, which is the case in our model. 

\section{Conclusion}
\label{sec:CraftNatural-Conclusion}
We develop a biologically motivated, feed-forward computational model of FGO 
with local and global cues. Spectral Anisotropy and T-Junctions are the local 
cues newly introduced into the model, which only influence the Orientation 
channel among the three feature channels. First, we show that even the reference 
model, with only a few global cues, convexity, surroundedness and parallelism, 
completely devoid of any local cues performs significantly better than chance 
level (50\%) achieving a FGCA of 58.44\% on the BSDS figure-ground dataset. Each 
local cue, when added alone leads to statistically significant improvement in 
the overall FGCA, compared to the reference model devoid of local cues, 
indicating their usefulness as independent local cues of FGO. The model with 
both SA and the T-Junctions achieves an 8.77\% improvement in terms of FGCA compared to 
that of the model without any local cues. Moreover, the FGCA of the model with both local cues 
is always higher than that of the models with individual local cues, indicating 
the mutually facilitatory nature of local cues. In conclusion, SA and 
T-Junctions are useful, mutually beneficiary local cues and lead to 
statistically significant improvement in the FGCA of the feed forward, 
biologically motivated FGO model, either when added alone or together. 

As we show in \ref{subsec:TJCompComplexity}, the computational complexity of 
adding 
both local cues is relatively low, yielding $\approx 9\%$ improvement in model's 
performance. Given that the feature channel weights are un-optimized, model 
consists of only a few global 
and local cues, local cues added to only one of three feature channels and the 
model is not optimized for best 
FGCA\footnote{See Chapter~\ref{sec:CraftNatural-FutureWork} for a discussion on 
how FGCA of 
the model can be improved even with existing local cues.}, the performance of 
the model is highly impressive.

\section{Future Work}
\label{sec:CraftNatural-FutureWork}

In future, we intend to improve the FGCA of the model by tuning the inhibitory 
weight, $w_{opp}$ for each feature and each 
local cue (Eqs.~\ref{eq:FGO-BOPyr-L} -- \ref{eq:FGO-BOPyr-TJ}) and tuning 
feature specific weights in Eq~\ref{eq:FGO-Final-BO-maps}. 
In addition, increasing the number of scales, having $\mathcal{CS}$ cells and 
$\mathcal{B}$ cells of multiple radii can all 
lead to better FGCA. $\mathcal{CS}$ cells $\mathcal{B}$ cells of multiple radii 
would capture the convexity and surroundedness cues better. Also, the model's figure-ground 
response is computed by modulating the activity of $\mathcal{C}_{\theta}$ cells, 
which are computed using Gabor filter kernels. The response of 
$\mathcal{C}_{\theta}$ cells may not always exactly coincide with human drawn 
boundaries in the ground-truth, with which we compare the model's response to 
calculate FGCA. Hence, averaging the BO response in a small $2 \times 2$ pixel 
neighborhood and then comparing that with the ground-truth FG labels could yield 
improved FGCA. In future, we would like to explore these ideas in order to 
improve FGCA. Moreover, color based 
cues~\cite{TROSCIANKO1991,zaidi_li_2006}, global cues such as 
symmetry~\cite{Ardila_etal11} and medial axis~\cite{Ardila_etal12} can be 
incorporated to improve the FGCA and make the model more robust.

In the biologically plausible SA computation~\ref{subsec:SAcomputation}, we used 
the Complex cell responses in all our computation. 
It would be interesting to see if similar or better FGCA can be achieved with 
Simple Even or Odd cells alone. In that case, the cost 
of computing SA would reduce by more than half. This would make the overall FGO 
model computation even more efficient. 
Also, in Section~\ref{subsec:SAcomputation}, filter size increment was in steps 
of 2 pixels. Having finer filter size resolutions (for example, $9 \times 9, 10 
\times 10, 
\ldots$ instead of $9 \times 9, 11 \times 11, \ldots$) will be considered to 
improve the FGCA even more. 

From a computational cost perspective, image segmentation using the 
gPb~\cite{amfm_pami2011} algorithm is the most expensive step in the FGO model 
with local cues. In order to decrease the computational cost, more efficient 
image segmentation algorithms should be explored. One efficient algorithm 
with similar performance as gPb (F-score, \citet{amfm_pami2011} = 0.70 \vs 
F-score, \citet{leordeanu2014generalized} = 0.69 on BSDS 500 dataset)  by 
\citet{leordeanu2014generalized}  is a good candidate. Replacing 
gPb~\cite{amfm_pami2011} algorithm with the algorithm by 
\citet{leordeanu2014generalized} for image segmentation, hence T-Junction 
computation, can substantially reduce the computational overload, while 
achieving similar performance. Other recent methods with better image 
segmentation performance can also be considered. We would also consider 
parallelization of the model using GPUs and FPGAs in future.

\section*{References}
\bibliographystyle{model1-num-names}
\bibliography{thesis,refs_old,ciss2011,ciss2013avsal,ciss2012}
\newpage

\appendix
\noindent{\bf \Large Supplementary Information}
\label{sec:supinfo}

\section{Computational complexity of adding local cues}
\label{subsec:TJCompComplexity}
The most computationally intensive part of SA computation is the correlations 
involved in Eq~\ref{eq:SACompPlus}, which 
has a computational complexity of $O(N_{r} \times N_{c} \times \log(N_{r} \times 
N_{c})) $ when implemented in Fourier domain, where $N_{r}$ and $N_{c}$ are the 
number of rows and columns in the image. 

The computationally intensive part of T-junction 
computation is the gPb~\cite{amfm_pami2011} based image segmentation. We utilize this algorithm \textit{as is}, hence we will not delve 
into exact estimation of computational complexity for this step. Once the 
contours and segmentation maps are obtained using gPb algorithm, the computation 
of each T-Junction using both methods described in 
Section~\ref{subsubsec:TJcompAreaBased} and 
Section~\ref{subsubsec:TJcompAngleBased} involves multiplying the edge maps, 
segmentation maps with masks of appropriate sizes, counting and tracking pixels, 
computing angles, \etc, which roughly translates into a computational complexity 
of $O(N_{mask})^2$ for both methods, where $N_{mask} = 13$ pixels for Segment 
Area based T-Junction computation (Section~\ref{subsubsec:TJcompAreaBased}) and 
$N_{mask} = 15$ pixels for Contour Angle based T-Junction computation 
(Section~\ref{subsubsec:TJcompAngleBased}). Typically $3-10$ T-Junctions are 
found in an image. So, once edges/segmentation map is computed, since only few 
T-Junctions are typically present in images and the size of mask is not very 
large, subsequent computation is not very time consuming. With appropriate modifications, it should be 
possible to reduce the computational complexity of T-Junction determination even further, 
which is not optimized at the moment.

\section{Local cues influencing only top 2 layers}
\label{app:2layerLocalCueFGCA}
Should the influence of local cues 
also be strictly local? Local cues, by definition, should be computed based on 
the analysis of a small patch of an image to determine figure-ground relations. 
This is what makes them computationally more efficient. But, should their 
influence also be local? There is no \emph{a priori} reason why their influence 
should be strictly local. To verify whether there is higher benefit in adding 
them locally only at the top layer (\ie, at native image resolution only), we 
added them only at the top layer.  For SA it resulted in a noticeable, but very 
small improvement. For T-Junctions, the change was barely noticeable. This could 
be due to extremely small size of von Mises filter kernels that we use ($R_{0} = 
2$ pixels) in comparison with the images size ($481 \times 321$ pixels). So, we 
added the local cues to the top two layers. For each local cue added separately, 
the optimal parameters of the model were recomputed and  those parameters were 
used to compute the FGCA. The versions of the model with local cues only at the 
top 2 layers did not give rise to better FGCA than what we saw earlier with the 
cues added at all scales. The results are summarized in 
Table~\ref{tab:SATJtop2layers}.

\begin{table}
\centering
\begin{tabular}{|l|l|l|}
\hline
\textbf{Model}                                                                             & \textbf{k = 2}   & \textbf{k = 10}  \\ \hline
Ref Model                                                                         & -       & 58.44\% \\ \hline
Ref + SA                                                                          & 62.42\% & 62.69\% \\ \hline
\begin{tabular}[c]{@{}l@{}}Ref + T-Junctions\\ (gPb edges)\end{tabular}           & 59.12\% & 59.48\% \\ \hline
\end{tabular}
\caption[Local cues only at the top 2 layers]{Local cues only at the top 2 
layers: By adding each local cue only at the top 2 layers ($k = 2$), we see the 
FGCA we obtain is much lower than having them at all levels ($k = 10$)}

\label{tab:SATJtop2layers}
\end{table}
\end{document}